\documentclass{article} % For LaTeX2e
\usepackage{iclr2026_conference,times}

\usepackage{amsmath,amsfonts,bm}

\def\eqref#1{equation~\ref{#1}}
\def\1{\bm{1}}

\def\vtheta{{\bm{\theta}}}

\DeclareMathAlphabet{\mathsfit}{\encodingdefault}{\sfdefault}{m}{sl}
\SetMathAlphabet{\mathsfit}{bold}{\encodingdefault}{\sfdefault}{bx}{n}

\def\sD{{\mathbb{D}}}
\newcommand{\E}{\mathbb{E}}

\newcommand{\KL}{D_{\mathrm{KL}}}

\usepackage{hyperref}
\usepackage{url}
\usepackage{amsthm}
\newtheorem{proposition}{Proposition}

\newtheorem{lemma}{Lemma}
\usepackage{graphicx}    % 插图
\usepackage{subcaption}  % 子图
\usepackage{makecell}
\usepackage{multirow}
\usepackage{pgfplots}
\pgfplotsset{compat=1.18}
\usepackage{wrapfig}   % 实现文字环绕图片的核心宏包
\usepackage{xcolor}
\usepackage{color}
\usepackage{colortbl}  % 用于表格颜色设置
\definecolor{mylightblue}{rgb}{0.9, 0.9, 1.0}

\title{Learning More with Less: A Dynamic Dual-Level Down-Sampling Framework for Efficient Policy Optimization
}

\iclrfinalcopy
\newcommand{\aspace}{\hspace{1em}}
\newcommand{\tsinghua}{$^{1}$}
\newcommand{\tencent}{$^{2}$}
\newcommand{\ucdavis}{$^{3}$}

\author{
  \parbox{\linewidth}{\centering
    Chao Wang\tsinghua \thanks{Work done while interning at Tencent.}\aspace
    Tao Yang\tencent\thanks{Corresponding Author.} \aspace
    Hongtao Tian\tencent \aspace
    Yunsheng Shi\tencent \aspace
    Qiyao Ma\ucdavis \footnotemark[1] \aspace
    Xiaotao Liu\tencent \aspace
    Ting Yao\tencent \aspace
    Wenbo Ding\tsinghua\footnotemark[2] 
  }
  \\
  \parbox{\linewidth}{\centering
  \tsinghua Tsinghua University \aspace
  \tencent WeChat, Tencent \aspace
  \ucdavis University of California, Davis \aspace
  }
}

\begin{document}

\maketitle

\begin{abstract}
%Reinforcement Learning from Human Feedback (RLHF) has emerged as a cornerstone for aligning Large Language Models (LLMs). However, mainstream policy optimization algorithms often grapple with efficiency bottlenecks. 
Critic-free methods like GRPO reduce memory demands by estimating advantages from multiple rollouts but tend to converge slowly, as critical learning signals are diluted by an abundance of uninformative samples and tokens. To tackle this challenge, we propose the \textbf{Dynamic Dual-Level Down-Sampling (D$^3$S)} framework that prioritizes the most informative samples and tokens across groups to improve the efficient of policy optimization. D$^3$S operates along two levels: (1) the sample-level, which selects a subset of rollouts to maximize advantage variance ($\text{Var}(A)$). We theoretically proven that this selection is positively correlated with the upper bound of the policy gradient norms, yielding higher policy gradients. (2) the token-level, which prioritizes tokens with a high product of advantage magnitude and policy entropy ($|A_{i,t}|\times H_{i,t}$), focusing updates on tokens where the policy is both uncertain and impactful. Moreover, to prevent overfitting to high-signal data, D$^3$S employs a dynamic down-sampling schedule inspired by curriculum learning. This schedule starts with aggressive down-sampling to accelerate early learning and gradually relaxes to promote robust generalization. Extensive experiments on Qwen2.5 and Llama3.1 demonstrate that integrating D$^3$S into advanced RL algorithms achieves state-of-the-art performance and generalization while requiring \textit{fewer} samples and tokens across diverse reasoning benchmarks. Our code is added in the supplementary materials and will be made publicly available.

\end{abstract}

\section{Introduction}

Reinforcement Learning (RL) has become instrumental in aligning Large Language Models (LLMs) with human values and preferences \citep{ouyang2022TrainingLanguageModels}, leading to the emergence of various alignment algorithms. 
%Among these, Proximal Policy Optimization (PPO) \citep{schulman2017ProximalPolicyOptimizationc} is notable for its strong performance but incurs significant computational costs, as it requires managing four large models: an actor, a critic, a reward model, and a reference model\citep{yu2025DAPOOpenSourceLLM}. To address this, 
Among these, critic-free methods like Group Relative Policy Optimization (GRPO) \citep{shao2024DeepSeekMathPushingLimits} and Group Sequence Policy Optimization (GSPO) \citep{zheng2025GroupSequencePolicy} have marked a crucial step towards greater memory efficiency. These methods estimate advantages using relative rewards from a group of sampled responses, thereby eliminating the need for a separate critic network \citep{shao2024DeepSeekMathPushingLimits}. However, while the memory bottleneck is alleviated, efficiency challenges remain. The precision of estimation of advantages utilized in training depends on the quality of the sampled groups. Larger groups risk diluting critical learning signals from a few key samples and tokens, as these signals can be overshadowed by the averaging effect of numerous \textit{undifferentiated} samples (e.g., groups dominated by mostly correct or incorrect samples in mathematical reasoning tasks). Conversely, smaller groups may struggle to yield diverse samples due to insufficient sampling. This trade-off constrains the optimization efficiency of critic-free algorithms.

%In recent years, Reinforcement Learning from Human Feedback (RLHF) has become a cornerstone for aligning Large Language Models (LLMs) with human values and preferences \citep{ouyang2022TrainingLanguageModels}. Proximal Policy Optimization (PPO) \citep{schulman2017ProximalPolicyOptimizationc}, with its robust performance, comes at a significant cost, requiring the simultaneous maintenance of four large models—an actor, a critic, a reward model, and a reference model\citep{yu2025DAPOOpenSourceLLM}. Critic-free methods like Group Relative Policy Optimization (GRPO) and Group Sequence Policy Optimization (GSPO) \citep{zheng2025GroupSequencePolicy} have marked a crucial step towards greater memory efficiency by estimating advantage signals from the relative rewards within a group of sampled responses, thereby eliminating the need for a separate critic network \citep{shao2024DeepSeekMathPushingLimits}. While the memory bottleneck is alleviated, another issue of efficiency persists. The precision of estimation of advantage signals utilized in training depends on the scale of sampling to build groups. When groups grow larger, the risk increases that the valuable learning signals from a few critical samples and tokens are averaged out by the near-zero or noisy signals from the vast majority of uninformative data points. This results in a smaller, less decisive gradient update step, slowing down convergence.

To tackle this issue, \citet{razin2024VanishingGradientsReinforcementa,razin2025WhatMakesReward} reveals that raising the variance of reward signals can accelerate convergence, as higher reward variance ($\text{Var}(R)$) creates a steeper optimization landscape. 
%Similarly, PODS \citep{xu2025NotAllRolloutsb} down samples a subset of rollouts with maximized reward variance ($\text{Var}(R)$) and utilizes them for GRPO. 
However, for typically critic-free methods (e.g., GRPO and GSPO), advantages are computed by normalizing the selected subset, resulting a fixed advantage variance of $1$.
Our theoretical analysis in Section~\ref{Section:Preliminaries} demonstrates that maximizing $\text{Var}(R)$ in critic-free methods imposes a \textit{fixed} upper bound on the policy gradient norm, whereas maximizing advantage variance ($\text{Var}(A)$) introduces a \textit{variable} upper bound positively correlated with the advantage variance. This indicates that maximizing $\text{Var}(A)$ has the potential to yield higher policy gradients within the core subset, thereby accelerating policy convergence toward the optimal trajectory. 

\begin{figure}[th]
\centering
\begin{subfigure}[b]{0.3\linewidth}
\includegraphics[width=\linewidth]{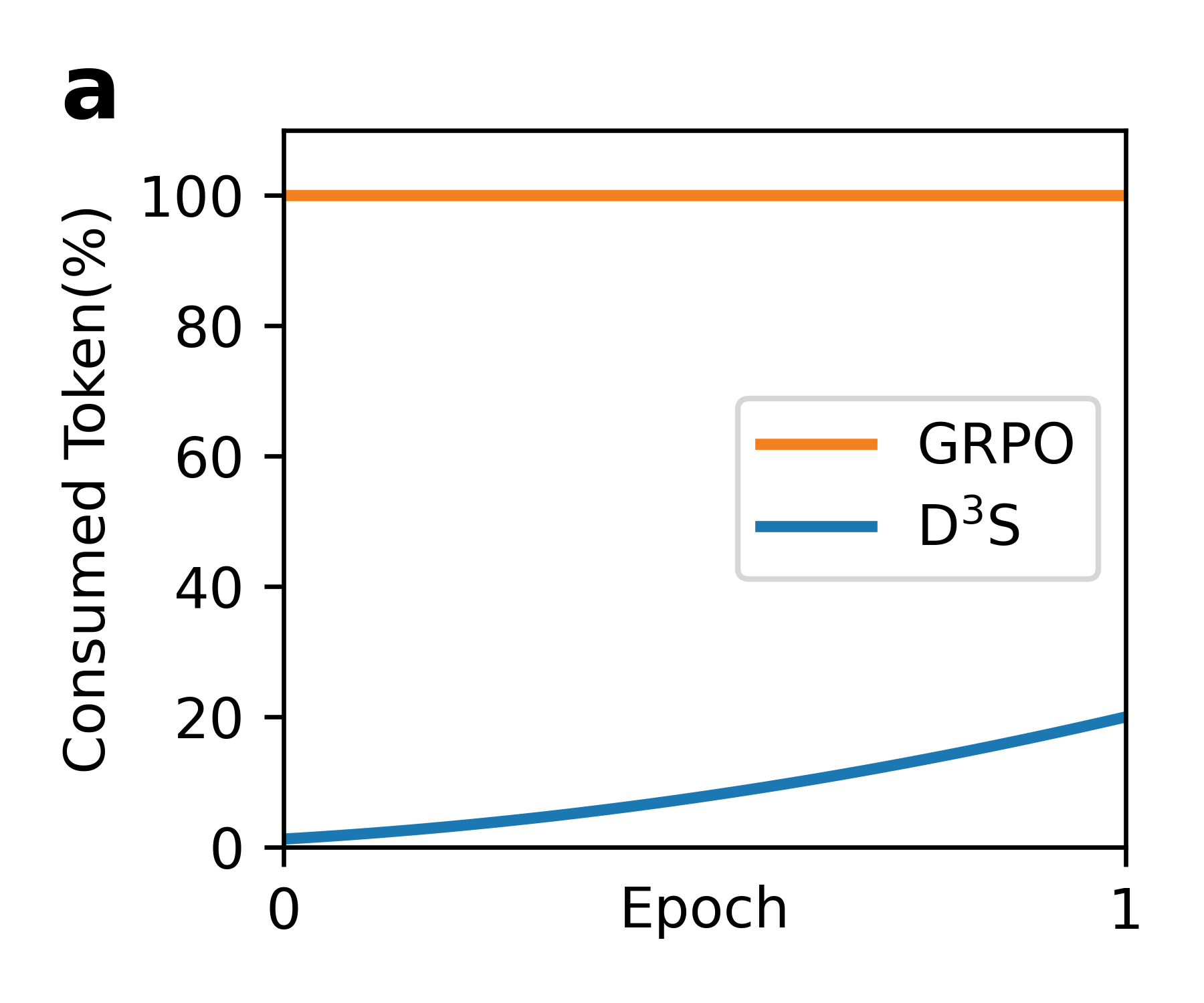}
\caption{}
\label{Figure:Train_Consumed_Token}
\end{subfigure}
%\hfill
\begin{subfigure}[b]{0.3\linewidth}
\includegraphics[width=\linewidth]{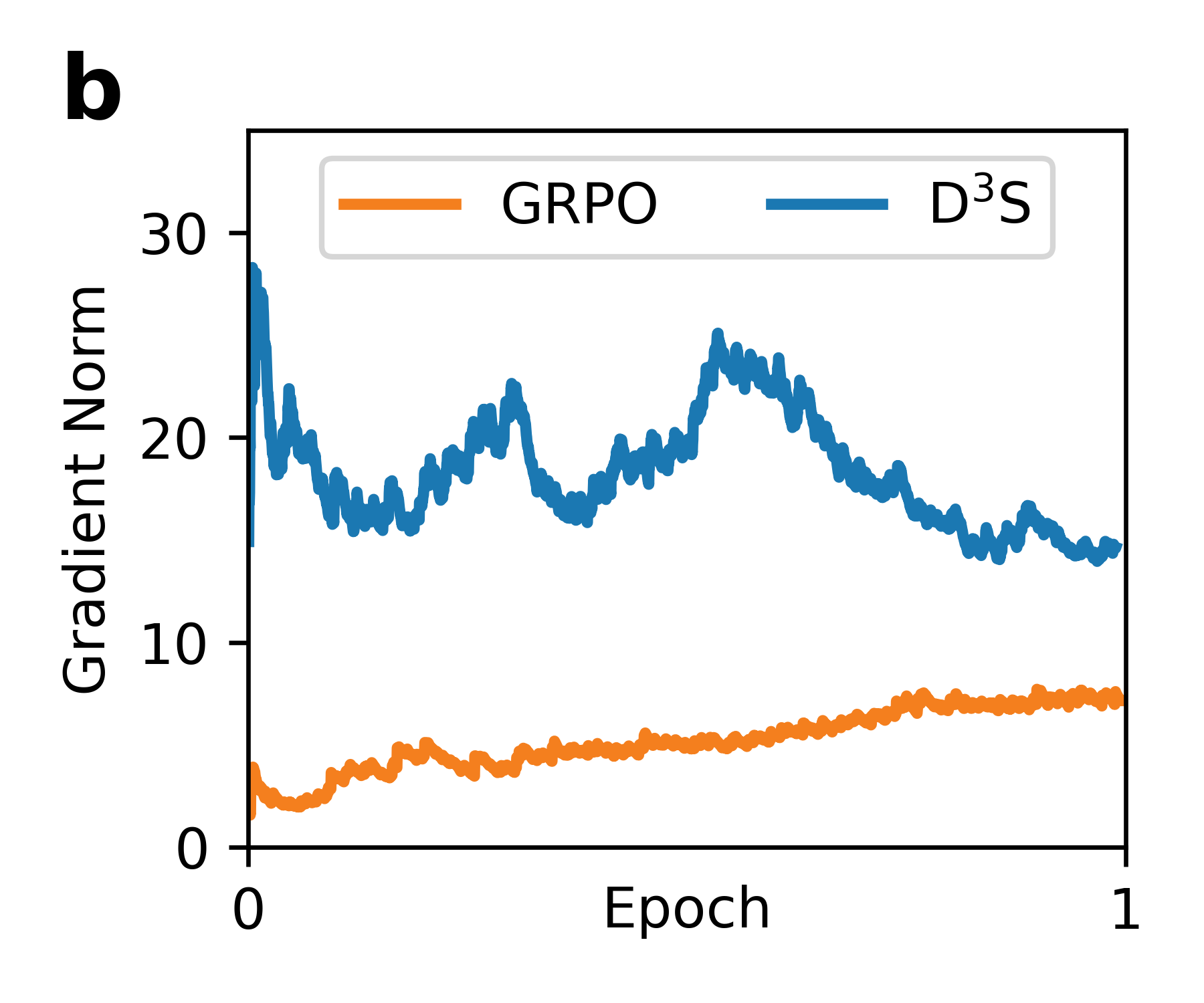}
\caption{}
\label{Figure:Train_Grad_Norm_Qwen_ema}
\end{subfigure}
%\hfill
\begin{subfigure}[b]{0.3\linewidth}
\includegraphics[width=\linewidth]
{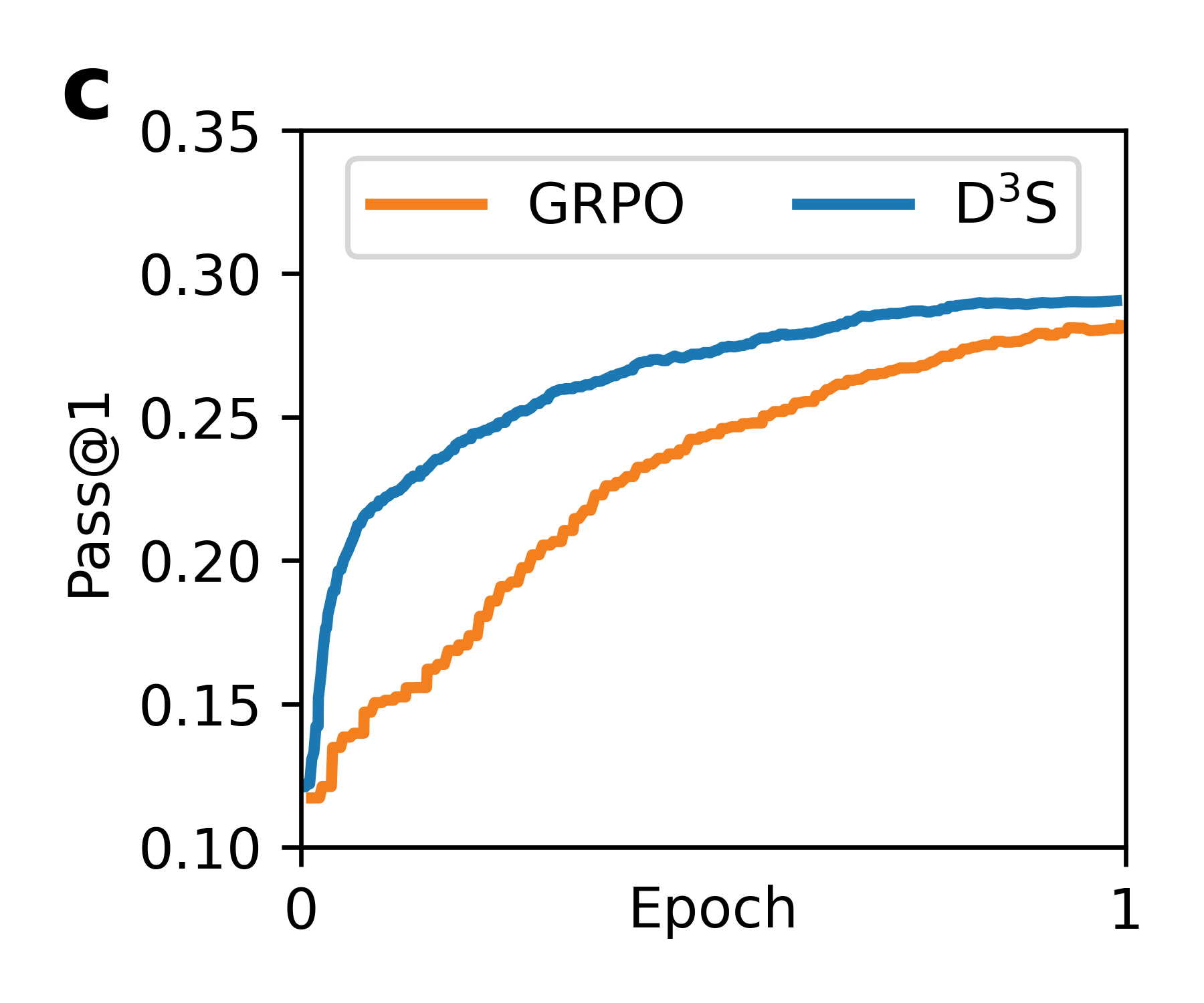}
\caption{}
\label{Figure:Train_Train_Qwen_Pass1_D3S_ema}
\end{subfigure}
\vspace{-10pt}
\caption{Comparison of training dynamics between D$^3$S and the GRPO: (\subref{Figure:Train_Consumed_Token}) token consumption ratio, (\subref{Figure:Train_Grad_Norm_Qwen_ema}) gradient norms, and (\subref{Figure:Train_Train_Qwen_Pass1_D3S_ema}) Pass@1 scores. Compared to the original GRPO, the integration of D$^3$S reduces token usage, accelerates policy convergence, and delivers superior performance.}
\label{Figure:Grad_Norm_vs_Pass1}
\end{figure}
% \vspace{-10mm}

Building on this insight, we propose \textbf{D}ynamic \textbf{D}ual-Level \textbf{D}own-\textbf{S}ampling (\textbf{D$^3$S}) framework, which operates on two levels. \textbf{First}, at the sample-level, instead of maximizing $\text{Var}(R)$, D$^3$S selects samples by first estimating the group-relative advantage across the entire batch and then maximizing $\text{Var}(A)$ to identify the core subset for optimization. 
\textbf{Second}, at the token-level, we further consider both advantage and entropy metrics, proposing the product of advantage magnitude and entropy ($|A_{i,t}| \times H_{i,t}$) as a measure of token importance. The advantage magnitude reflects the relative significance of tokens, while entropy quantifies uncertainty—an essential factor for guiding reasoning paths in reasoning tasks \citep{wang20258020Rule}. Moreover, to prevent the policy from overfitting to a limited set of high-signal data and compromising its generalization, we introduce a dynamic down-sampling schedule. Inspired by curriculum learning \citep{bengio2009CurriculumLearning}, which progresses from simple to complex, the dynamic down-sampling schedule begins by prioritizing a smaller subset of high-signal samples and tokens to accelerate early-stage learning. As training progresses, the data pool gradually expanded, incorporating more samples and tokens to improve generalization.

To verify the effectiveness of D$^3$S, we conduct extensive experiments across various RL settings (i.e., GRPO and GSPO) on challenging mathematical reasoning tasks. The results demonstrate that incorporating D$^3$S into GRPO and GSPO consistently improves performance while reducing sample and token requirements. A direct comparison of training dynamics, as illustrated in Figure~\ref{Figure:Grad_Norm_vs_Pass1}, uses Qwen2.5-Math-7B \citep{yang2024qwen25mathtechnicalreportmathematical} as the backbone and is trained on AIME24 \citep{MathaiAime24Datasets}. Compared to the original GRPO, D$^3$S optimizes fewer than 20\% of the tokens (Figure~\ref{Figure:Train_Consumed_Token}) while achieving higher policy gradients (Figure~\ref{Figure:Train_Grad_Norm_Qwen_ema}), leading to significantly faster convergence and superior Pass@1 scores on the test set (Figure~\ref{Figure:Train_Train_Qwen_Pass1_D3S_ema}). Specifically, when using Qwen2.5-Math-7B as the backbone, GRPO with D$^3$S achieves average improvements of 4.5 in Pass@1 and 3.7 in Pass@8 across seven datasets, compared to the original GRPO. Similarly, with Llama3.1-8B-Instruct \citep{grattafioriLlama3Herd2024} as the backbone, GRPO with D$^3$S outperforms the original by 3.3 in Pass@1 and 7.8 in Pass@8. Moreover, our analysis highlights the following key findings: (1) Both sample-level and token-level down-sampling effectively eliminate undifferentiated signals in the early training stages, accelerating policy convergence. (2) In the later training stages, the dynamic down-sampling schedule plays a crucial role in enhancing the generalization of D$^3$S. (3)  D$^3$S better manages entropy fluctuations, reflecting more stable policy training.

%We demonstrate the effectiveness of D$^3$S through post-training experiments with Qwen2.5-Math-7B, Llama3.1-8B-Instruct and OpenMath2-Llama3.1-8B (Llama3.1-8B-Instruct finetuned for math tasks by NVIDIA) as base models. As briefed in Figure~\ref{Figure:Grad_Norm_vs_Pass1}, D$^3$S is capable of generate policy gradients with larger norm and improve convergence while consuming far less tokens than GRPO. Further, as detailed in Section~\ref{Section:Experiment}, both GRPO \citep{shao2024DeepSeekMathPushingLimits} and GSPO \citep{zheng2025GroupSequencePolicy} base algorithms can be improved by D$^3$S to provide state-of-the-art performance on various benchmark tasks. The abilities of D$^3$S to focus on valuable data and dynamically schedule learning criteria are proved to be effective for group-relative advantage-based policy optimization algorithms.

% \vspace{-10mm}
\section{Theoretical Analysis}
\label{Section:upper_bound}
\subsection{Preliminaries}
\label{Section:Preliminaries}

% \paragraph{Notation}

% \paragraph{Group-Relative Advantage Estimation}
Formally, let $x$ represent an input query from the dataset $\sD$. A large language model with parameters $\vtheta$ is defined as a policy $\pi_{\vtheta}$. For each query, the policy $\pi_{\vtheta}$ generates multiple responses, with a group size of $G$. For the $i$-th response $y_i$, the number of tokens is denoted as ${|y_i|}$. GRPO \citep{shao2024DeepSeekMathPushingLimits} removes the need for a standalone value network by estimating group-relative advantages directly from $G$. The optimization objective is given as:
%Group Relative Policy Optimization (GRPO)~\citep{shao2024DeepSeekMathPushingLimits} use group-relative advantages instead of standalone value network. Without value network, for each queries $x$ in dataset $\sD$, GRPO self-iteratively updates policy $\pi_{\vtheta}$ utilizing estimation among $G$ responses $y_{i},i\in[1,G]$ sampled by policy $\pi_{\vtheta_{old}}$ of last optimization step. Formally, GRPO optimizes the following objective:
\vspace{0.1cm}
\begin{align}
J_{\text{GRPO}}(\vtheta) &= \E_{x \sim \sD, \{y_i\}_{i=1}^G \sim \pi_{\vtheta_{\text{old}}}(\cdot|x)}  \\ 
& \left[
\frac{1}{G} \sum_{i=1}^G \frac{1}{|y_i|} \sum_{t=1}^{|y_i|} \min(w_{i,t}(\vtheta) A_{i,t}, \text{clip}(w_{i,t}(\vtheta), 1-\varepsilon, 1+\varepsilon) A_{i,t})  - \beta \KL ( \pi_{\vtheta} \Vert \pi_{ref} )
\right] \nonumber
\end{align}

where the importance ratio is $w_{i,t}(\vtheta) = \frac{\pi_\vtheta(y_{i,t}|x, y_{i,<t})}{\pi_{\vtheta_{\text{old}}}(y_{i,t}|x, y_{i,<t})}$ which will be clipped by hyper-parameter $\epsilon$. $\beta$ regulates the constraint on the KL-divergence $\KL$. The group-normalized advantage is derived by standardizing the reward signal $R$ within each group:
\vspace{0.1cm}
\begin{align}
A_i = \frac{R(x, y_i) - \frac{1}{G}\sum_{j=i}^G R
(x, y_i)}{\text{std}(\{R(x, y_i)\}_{i=1}^G)}
\label{Equation:Advantage}
\end{align}

Building upon GRPO, \citet{zheng2025GroupSequencePolicy} proposes GSPO, which utilizes the full sequence as context for next-token prediction. While adopting the same token-level group-relative advantage signal defined in Equation~\ref{Equation:Advantage}, GSPO further incorporates a novel importance ratio based on sequence likelihood. The optimization objective of GSPO is denoted as:
%Based on GRPO, \citet{zheng2025GroupSequencePolicy} introduce GSPO which takes whole sequence as context of next-token-prediction. Besides using identical token-level group-relative advantage signal defined in Equation~\ref{Equation:Advantage}, GSPO introduces a new importance ratio based on sequence likelihood. Formally, the optimization objective of GSPO is:
\vspace{0.1cm}
\begin{align}
J_{\text{GSPO}}(\vtheta) &= \E_{x \sim \sD, \{y_i\}_{i=1}^G \sim \pi_{\vtheta_{\text{old}}}(\cdot|x)} \\ 
& \left[
\frac{1}{G} \sum_{i=1}^G \frac{1}{|y_i|} \sum_{t=1}^{|y_i|} \min(s_{i,t}(\vtheta) A_{i,t}, \text{clip}(s_{i,t}(\vtheta), 1-\varepsilon, 1+\varepsilon) A_{i,t})  - \beta \KL ( \pi_{\vtheta} \Vert \pi_{ref} )
\right]  \nonumber
\end{align}
where the importance ratio is $s_{i,t}(\vtheta) = \text{sg}\left[\left(\frac{\pi_{\vtheta}(y_{i}|x)}{\pi_{\vtheta_{\text{old}}(y_{i}|x)}}\right)^{\frac{1}{|y_{i}|}}\right] 
\cdot 
\frac{\pi_\vtheta(y_{i,t}|x, y_{i,<t})}{\text{sg}[\pi_{\vtheta_{\text{old}}}(y_{i,t}|x, y_{i,<t})]}$ with $\text{sg}[\cdot]$ denoting stopping gradient. The group-normalized advantage is same as Equation~\ref{Equation:Advantage}.
% \subsection{Improvement }

\subsection{Upper Bounds on Gradient Norms}
\label{Section:upper_bond}

We begin by analyzing the upper bound of the gradient norm in GRPO (also applicable to GSPO), as it determines the scale of policy updates and directly impacts training efficiency. 

\begin{proposition}\label{Proposition:J_GRPO}
    
The gradient of $J_{\text{GRPO}}$ satisfies:
\vspace{0.1cm}
\begin{align}
\|\nabla_\vtheta J_{\text{GRPO}}(\vtheta)\| \leq 4\gamma(x;\vtheta)
\end{align}

where $\gamma(x;\vtheta)$ denotes a static parameter related to input $x$ and model parameters $\vtheta$.
\end{proposition}
\vspace{-0.4cm}
\begin{proof}
    See Appendix \ref{Section:Proof_Proposition_1}.
\end{proof}
\vspace{-0.3cm}

Proposition \ref{Proposition:J_GRPO} highlights a key property of group-based methods: the gradient norm is capped by a fixed upper bound, regardless of the explicit reward variance $\text{Var}(R)$. We then extend Proposition \ref{Proposition:J_GRPO} to scenarios where optimization is performed on a subset sampled from the group $G$.

\begin{proposition}\label{Proposition:J_GRPO_subset}

The gradient of $\hat{J}_{\text{GRPO}}$ which selects a subset from original rollout group with size $G$ satisfies:
\begin{align}
\|\nabla_\vtheta \hat{J}_{\text{GRPO}}(\vtheta)\| \leq 3\cdot 2^{\frac{1}{3}} \cdot \gamma(x;\vtheta) \cdot (\sqrt{G - 1})^{1/3} \cdot (\text{Var}(A'))^{1/3}
\end{align}

where $\text{Var}(A')$ is the advantage variance of the selected subset from rollouts.
%and $\gamma(x;\vtheta)$ denotes a static parameter related to input query $x$ and model parameters $\vtheta$.
\end{proposition}
\vspace{-0.4cm}
\begin{proof}
    See Appendix \ref{Section:Proof_Proposition_2}.
\end{proof}
\vspace{-0.3cm}

\begin{lemma}
\label{Lemma:Max_Var}
Let $A$ be a set of $M$ elements that is standardized as $\E[A]=0, \operatorname{Var}(A) = 1$. For any integer $N$ such that $2 \le N \le M$, there exists a subset $A' \subseteq A$ with $|A'|=N$ satisfying $\operatorname{Var}(A') \ge 1$.
\end{lemma}
\vspace{-0.4cm}
\begin{proof}
    See Appendix \ref{Section:Proof_Lemma_1}.
\end{proof}
\vspace{-0.3cm}

From Proposition~\ref{Proposition:J_GRPO_subset}, we can observe that under the down-sampling perspective, the gradient norm of GRPO has a \textit{variable} upper bound, which is positively correlated with the advantage variance $\text{Var}(A')$ of the subset. Consequently, the strategy of maximizing $\text{Var}(R)$ \citep{razin2024VanishingGradientsReinforcementa, razin2025WhatMakesReward,xu2025NotAllRolloutsb} in group-based methods encounters two major limitations. \textbf{First}, even if $\text{Var}(R)$ is maximized within the selected subset, the advantages are still computed under the constraint of a fixed $\text{Var}(A') = 1$ and cannot change the gradient norm upper bound. \textbf{Second}, the advantages are estimated within a smaller, biased subset of the original group, leading to unstable advantage estimation. This naturally motivates an alternative approach: compute normalized advantages using all samples from the original group, and then select a subset that maximizes the variance of these normalized advantages. Lemma~\ref{Lemma:Max_Var} proves that a subset with a variance of at least 1 can be extracted from a set of normalized advantages. This indicates that such a subset leads to a higher upper bound on the gradient norm. The trends in Figure~\ref{Figure:Train_Grad_Norm_Qwen_ema} provide empirical evidence supporting this property.

\section{Method}\label{Section:Method}

In this study, we introduce the \textbf{Dynamic Dual-level Down-Sampling (D$^3$S)} framework, which improves training efficiency through a two-tier approach: a sample-level down-sampling strategy and a token-level selection mechanism.

% While GRPO provides a stable foundation for RL training, we observe that computational efficiency can be significantly improved through strategic sampling without compromising convergence guarantees. 
% , which introduces sample and token dual-level down sampling and dynamic learning schedule to amplify policy gradient magnitudes while maintaining distributional properties.

% \subsection{Improve Exploitation}\label{Section:Improve_Exploitation}
\subsection{Sample-level: Cross-Group Advantage-Based Down-Sampling}\label{Section:Sample_level}

Building on the insights from Section \ref{Section:upper_bond}, rather than selecting subsets of rollouts based on maximizing $\text{Var}(R)$ \citep{razin2024VanishingGradientsReinforcementa, razin2025WhatMakesReward,xu2025NotAllRolloutsb}, we propose a refined method that identifies a core subset of samples based on their group-relative estimated advantages. This approach prioritizes maximizing the variance of advantage signals within the selected subset. Formally, given a query $x$ and its rollouts $\mathcal{S}_{\text{query}} = \{(x, y_{i}, A_{i}) : i \in [1,G]\}$, the selected subset $\hat{\mathcal{S}}_{\text{query}}$ is defined as follows:
\vspace{0.1cm}
\begin{equation}
\begin{gathered}
\hat{\mathcal{S}}_{\text{query}} 
=  \underset{\hat{S} \subset \mathcal{S}_{\text{query}},\, |\hat{S}|=N_{\hat{s}}}{\arg\max} \operatorname{Var}(A_{\hat{S}}) 
% = \mathrm{top}_{N_s}\big(\mathcal{S}_{\text{query}},\, \text{key}=|A_{i}|\big)
\end{gathered}\label{Equation:Max_Var_A_inner}
\end{equation} 

%We propose \textbf{Dynamic Dual-level Down Sampling (D$^3$S)} framework. Based on Proposition~\ref{Proposition:J_GRPO} and ~\ref{Proposition:J_GRPO_subset}, we introduce sample and token dual-level down sampling and dynamic learning schedule to amplify policy gradient magnitudes of GRPO. 
% Consequently, to accelerate convergence, one must increase the variance of the selected set of samples, which directly enlarges the attainable gradient magnitude. 
%This perspective resonates with recent works emphasizing the role of variance of reward signal \citep{razin2025WhatMakesReward,razin2024VanishingGradientsReinforcementa} . D$^3$S first select a core subset of samples according to their group-relatively estimated advantages and maximizing the variance of advantage signals within the subset. Formally, given a query $x$ and its $G$ rollouts $\mathcal{S}_{\text{query}} = \left\{(x, y_{i}, A_{i})\right\}$, instead of using all $G$ responses, we select a subset with maximized advantage variance
% \begin{equation}
% \begin{gathered}
% \mathcal{S}_{\text{query}} = \{(x, y_{i}, A_{i}) : i \in [1,G]\}, \ \ \ 
% \hat{\mathcal{S}}_{\text{query}} 
% =  \underset{S \subset \mathcal{S}_{\text{query}},\, |S|=N_s}{\arg\max} \operatorname{Var}(A_S) 
% % = \mathrm{top}_{N_s}\big(\mathcal{S}_{\text{query}},\, \text{key}=|A_{i}|\big)
% \end{gathered}\label{Equation:Max_Var_A_inner}
% \end{equation} 
where $N_{\hat{s}}$ is the number of selected samples within the group $\mathcal{S}_{\text{query}}$, and $A_{\hat{S}}=\{A_1, A_2,...,A_{|\hat{S}|}\} $ denotes the advantage set of $\hat{\mathcal{S}}_{\text{query}}$. Each advantage ${A}_i$ is defined as the mean of the token-level advantages ${A}_{i,t}$ over the sequence. \citet{xu2025NotAllRolloutsb} shows that the subset maximizing variance can be efficiently selected from the \( \tbinom{N_{\hat{S}}}{G} \) possible combinations, where \( N_{\hat{S}} = N_{{\hat{S}},\text{pos}} + N_{{\hat{S}},\text{neg}} \). In our scenario, we adopt this implementation, where \( N_{{\hat{S}},\text{pos}} \) refers to the samples with the highest positive advantages, while \( N_{{\hat{S}},\text{neg}} \) corresponds to those with the lowest negative advantages. 

Moreover, considering that gradient updates are performed in batches and there are significant advantage disparities across different groups (e.g., groups with all correct or incorrect predictions may result in uninformative zero advantages), we introduce a cross-group operation to select a high-variance subset from the entire batch. Formally, let $\mathcal{S}_{\text{batch}} = \{\mathcal{S}_{\text{query},b}: b \in [1,B]\}$ be a batch of rollouts, where $B$ is the batch size and $N=B\times N_{\hat{s}}$ is the total number of selected samples. The high-variance subset can then be obtained as:
\vspace{0.1cm}
\begin{equation}
\begin{gathered}
\hat{\mathcal{S}}_{\text{batch}} 
=  \underset{\hat{S} \subset S_{\text{batch}}, |\hat{S}|=N}{\arg\max} \operatorname{Var}(A_{\hat{S}}) \ 
% = \text{top}_{N}(\mathcal{S}_{\text{batch}}, \text{key}=|A_{i}|)
\end{gathered}\label{Equation:Max_Var_A_cross}
\end{equation}

The cross-group operation retains the original distributional properties, as the advantages are pre-normalized within each group, and no additional normalization is applied.

\subsection{Token-level: Entropy-Advantage Weighted Selection}\label{Section:Token_level}

Intuitively, responses often consist of a combination of easy tokens (where the model is both confident and accurate), neutral tokens (with minimal impact on outcomes), and critical tokens (where the model is uncertain and decisions significantly influence rewards). Policy entropy serves as a measure of the model's uncertainty \citep{cui2025EntropyMechanismReinforcement} and as an indicator of potential performance gains. \citet{wang20258020Rule} demonstrates that the top 20\% of tokens with the highest entropy dominate the policy gradient. Treating all tokens equally during updates dilutes the gradient signal, akin to averaging out meaningful information with noise.

To this end, we propose a token-level selection mechanism that integrates generation entropy and its advantage into a unified importance metric for ranking tokens across all selected samples. Specifically, high-importance tokens are identified as follows:
\begin{equation}
\begin{gathered} \label{Equation:Entropy}
H_{{i,t}}
= -\sum_{j=1}^{V} \pi_\theta(\text{token}_j \mid x_{i}, y_{i,<t}) \log \pi_\theta(\text{token}_j \mid x_{i}, y_{i,<t}) \\
\mathcal{T} = \text{top}_{K\%}(y_{i,t}, y_{i,t} \in \hat{\mathcal{S}},\text{key}=|A_{{i,t}}|\times H_{{i,t}})
\end{gathered}
\end{equation}

where $H_{{i,t}}$ denotes the entropy of $t$-th token in $i$-th response, and $K$ indicates the proportion of selected tokens. High entropy $H_{i,t}$ indicates higher uncertainty in the model's decision at that token position, fostering exploration and enhancing policy diversity. Advantage $A_{i,t}$ quantifies the impact of token on policy improvement, where a larger $|A_{i,t}|$ reflecting greater potential—positive or negative—for optimization. During each update, only the top \( K\% \) of tokens contribute to the gradient of \( \vtheta \). By selecting tokens with high entropy and advantage, computation is focused on the most informative decision points, encouraging the policy to resolve uncertainty in reward-critical regions.

\subsection{Dynamic Down-Sampling Schedule}
\label{Section:Dynamic_Sampling_Schedule}

The sample-level and token-level strategies discussed in Sections~\ref{Section:Sample_level} and~\ref{Section:Token_level} refine policy updates by prioritizing high-signal samples and tokens. Although this approach accelerates convergence by leveraging stronger gradients, it also heightens the risk of reward hacking or overfitting. The model might over-exploit a limited set of trajectories that seem highly informative in the early stages, but struggle to generalize as optimization progresses. To address this, we introduce a dynamic down-sampling schedule that progressively reduces the intensity of down-sampling as training advances.

Specifically, we employ a linear schedule to interpolate between the initial aggressive configuration \((N_{\text{init}}, K_{\text{init}})\) and the final milder configuration \((N_{\text{final}}, K_{\text{final}})\), based on the training progress \(p \in [0, 1]\). We use \( N_s^{(p)} \) to regulate the number of retained samples per query, while \( K^{(p)} \) determines the proportion of retained tokens within each sample:
\vspace{0.1cm}
\begin{align}
\left[N_s^{(p)},K^{(p)} \right] &= (1-p)\cdot \left[N_{\text{init}},K_{\text{init}} \right]+ p \cdot \left[N_{\text{final}},K_{\text{final}} \right] 
\label{Equation:Dynamic_Schedule}
\end{align}

At the beginning of training ($p = 0$), the model prioritizes the most informative rollouts and tokens to accelerate learning. As training advances ($p \to 1$), its focus broadens, incorporating more diverse signals to enrich learning and expand its scope. As empirical results shown in Figure~\ref{Figure:Dynamic_vs_Static_Qwen}, variance-based selection enhances early performance but loses effectiveness over time and risks overfitting. In contrast, the dynamic schedule sustains both a fast convergence rate and consistent improvements.

\subsection{\texorpdfstring{D$^3$S}{D3S} Optimization Objective}

By integrating the sampling strategies outlined above, the objective function of D$^3$S is expressed in Equation~\ref{Equation:J_D3S}.
%Combining sampling strategies defined above, the objective function of D$^3$S is formulated as Equation~\ref{Equation:J_D3S}. D$^3$S applies selection masks upon GRPO, rather than either modify original advantage estimation or remove $\KL$ restrain, maintaining the original convergence capability of GRPO.
\vspace{0.1cm}
\begin{align}
J_{\text{D$^3$S}}(\vtheta) &= \E_{x \sim \sD, \{y_i\}_{i=1}^G \sim \pi_{\vtheta_{\text{old}}}(\cdot|x)} \nonumber \\
&
\frac{1}{|\hat{\mathcal{S}}|} \frac{1}{|\mathcal{T}|}\sum_{i=1}^{i \in \hat{\mathcal{S}}}  \sum_{t=1}^{t \in \mathcal{T}} \left\{ \min \left[ w_{\vtheta,i,t} A_{i,t}, \text{clip}(w_{\vtheta,i,t}, 1-\varepsilon, 1+\varepsilon) A_{i,t} \right] - \beta \KL ( \pi_{\vtheta} \Vert \pi_{ref} )
\right\}
\label{Equation:J_D3S}
\end{align}

\section{Experiment}
\label{Section:Experiment}
\subsection{Configuration}
\label{Section:Experiment_Configuration}
\paragraph{Datasets and Evaluation}
We train each model using the DeepScaleR \citep{deepscaler2025} dataset, which includes AIME problems from 1984 to 2023, AMC problems prior to 2023, and questions from other sources 
%such as Omni-MATH \citep{gao2024OmniMATHUniversalOlympiad} and Still
. During training, the AIME24 \citep{MathaiAime24Datasets} is used as a validation set to monitor the out-of-domain performance of policy in real time. Evaluation is conducted on a diverse set of benchmarks, including AIME25 \citep{MathaiAime25Datasets}, AIME24 \citep{MathaiAime24Datasets}, AMC23 \citep{MathaiAmc23Datasets}, GSM8K \citep{cobbe2021TrainingVerifiersSolve}, MATH \citep{hendrycks2021MeasuringMathematicalProblema}, MinervaMath \citep{lewkowycz2022SolvingQuantitativeReasoning} and OlympiadBench \citep{he2024OlympiadBenchChallengingBenchmark}. For each question in these benchmarks, we generate 32 parallel outputs and compute Pass@k metrics. Rewards are assigned to the entire sequence based on the correctness of the answer, verified using math\_verify tool \citep{kydlicek2025MathVerifyMathVerification}.

%In mathematical tasks, rewards are assigned to the entire sequence based on the correctness of the final answer, verified using math\_verify tool \citep{kydlicek2025MathVerifyMathVerification}. The advantage is derived from sample-level rewards using the group normalization defined in Equation~\ref{Equation:Advantage}.

%We finetune each model on DeepScaleR \citep{deepscaler2025} dataset. DeepScalR contains AIME problems from 1984-2023 and AMC problems prior to 2023, along with questions from other datasets including Omni-MATH \citep{gao2024OmniMATHUniversalOlympiad} and Still. During training, AIME24 \citep{MathaiAime24Datasets} benchmark is borrowed as validation dataset in order to measure the runtime dynamic of model's out-of-domain performance. Evaluation is conducted across various benchmark datasets, including AIME25 \citep{MathaiAime25Datasets}, AIME24 \citep{MathaiAime24Datasets}, AMC23 \citep{MathaiAmc23Datasets}, GSM8K \citep{cobbe2021TrainingVerifiersSolve}, MATH \citep{hendrycks2021MeasuringMathematicalProblema}, MinervaMath \citep{lewkowycz2022SolvingQuantitativeReasoning} and OlympiadBench \citep{he2024OlympiadBenchChallengingBenchmark}. For each question in each evaluation benchmark, we rollout 32 parallel runs and collect pass@k metrics. In math experiments, rewards are assigned to the entire sequence based on the correctness of the final answer using math\_verify \citep{kydlicek2025MathVerifyMathVerification}. The advantage is derived from sample-level rewards using the group normalization defined in Equation~\ref{Equation:Advantage}.

\paragraph{Models}
We employ various models to systematically evaluate the proposed D$^3$S framework. Qwen2.5-Math-7B \citep{yang2024qwen25mathtechnicalreportmathematical}, a pre-aligned model, is specifically optimized for mathematical tasks. Llama3.1-8B-Instruct \citep{grattafioriLlama3Herd2024} serves as a general-purpose baseline model, while OpenMath2-Llama3.1-8B \citep{toshniwal2024openmath2}, a fine-tuned variant of Llama3.1-8B-Instruct, is included for comparison. Besides, a smaller model, Qwen2.5-Math-1.5B\citep{yang2024qwen25mathtechnicalreportmathematical}, is utilized to assess adaptability across varying model scales. Each model is configured using its officially recommended settings as shown in Table~\ref{Table:Model_hyper_parameters}. 

\paragraph{Baselines}
To assess the effectiveness of our approach, we integrate D$^3$S into two popular RL algorithms: GRPO \citep{shao2024DeepSeekMathPushingLimits} and its variant, GSPO \citep{zheng2025GroupSequencePolicy}, which enhances GRPO by improving sequence-level advantage estimation. Additionally, we compare our method with PODS \citep{xu2025NotAllRolloutsb}, a down-sampling strategy via maximizing reward variance. To further analyze the impact of different stages of D$^3$S, we compare several variants: (1) D$^1$S, which applies sample-level down-sampling by maximizing advantage variance as defined in Equation~\ref{Equation:Max_Var_A_inner}; (2) D$^1$S w/\textbf{C}ross, which incorporates cross-group operation as described in Equation~\ref{Equation:Max_Var_A_cross}; and (3) D$^2$S, which combines sample-level and token-level down-sampling but excludes the use of the dynamic down-sampling schedule, as defined in Equation~\ref{Equation:Entropy}. Parameters are listed in Table~\ref{Table:Algorithm_hyper_parameters}.

\subsection{Main Results}

\begin{table}[ht]
\centering
\caption{Experimental results on various mathematical reasoning benchmarks using different model backbones. We report Pass@1/Pass@8 scores, computed from 32 parallel runs. The best results are highlighted in \textbf{bold}, while the second-best are \underline{underlined}.}
\label{Table:Pass_1_8}
\setlength{\tabcolsep}{3pt}
\setlength{\extrarowheight}{2pt}
\resizebox{0.8\textwidth}{!}{
\begin{tabular}{lcccccccc}
\hline
\textbf{Model}            & \textbf{AIME24} & \textbf{AIME25} & \textbf{AMC23} & \textbf{GSM8k} & \textbf{MATH} & \textbf{Minerva} & \textbf{Olympiad} & \textbf{Average} \\ \hline
\multicolumn{9}{c}{Qwen2.5-Math-7B} \\ \hline
Base &
  8.9/33.2 &
  2.3/13.4 &
  22.8/70.4 &
  30.1/83.2 &
  27.9/64.6 &
  8.4/33.7 &
  4.1/14.6 &
  14.9/44.7 \\
GRPO &
  13.2/37.6 &
  5.5/21.6 &
  47.0/\underline{83.5} &
  64.9/94.3 &
  48.5/70.2 &
  19.8/45.0 &
  9.7/19.8 &
  29.8/53.1 \\
\ \  +PODS &
  \underline{16.1}/\underline{40.5} &
  \underline{7.8}/\underline{24.5} &
  \underline{52.8}/81.5 &
  \underline{73.3}/\underline{95.0} &
  \textbf{53.0}/\underline{71.1} &
  \underline{24.6}/\underline{47.5} &
  \textbf{11.0}/\underline{20.7} &
  \underline{34.1}/\underline{54.4} \\
\rowcolor{mylightblue}\ \  +\textbf{D$^3$S} &
  \textbf{20.3/48.2} &
  \textbf{7.9/25.8} &
  \textbf{54.4/87.1} &
  \textbf{73.4}/\textbf{95.7} &
  \underline{52.2}/\textbf{71.5} &
  \textbf{25.0}/\textbf{48.2} &
  \underline{10.7}/\textbf{20.8} &
  \textbf{34.3/56.8} \\ \hline
GSPO                 & \underline{15.8}/\underline{42.4} & \underline{6.7}/\underline{25.3} & 50.8/81.2 & 72.0/95.2 & 52.1/71.0 & 24.2/47.3 & 10.8/20.7 & 33.2/\underline{54.7}        \\
\ \  +PODS                 & 15.4/40.9 & 6.5/22.9 & \underline{51.9}/\underline{81.6} & \underline{72.9}/\underline{95.3} & \underline{52.9}/\underline{71.1} & \underline{25.0}/\underline{47.6} & \underline{10.9}/\underline{20.9} & \underline{33.6}/54.3        \\
\rowcolor{mylightblue}\ \  +\textbf{D$^3$S} &
  \textbf{18.3/43.3} &
  \textbf{8.3/26.9} &
  \textbf{53.2/83.8} &
  \textbf{76.0/96.1} &
  \textbf{54.9/71.4} &
  \textbf{28.4/51.1} &
  \textbf{11.5/21.1} &
  \textbf{35.8/56.2} \\ \hline
\multicolumn{9}{c}{Qwen2.5-Math-1.5B}                                                                  \\ \hline
Base    & 4.7/23.7  & 2.1/13.2 & 21.3/60.8 & 23.7/75.8 & 18.6/56.3 & 7.5/30.0  & 5.0/15.8  & 11.8/39.4 \\
GRPO    & 10.0/28.3 & \underline{6.1}/19.9 & 46.2/\underline{77.0} & \underline{77.3}/\textbf{94.4} & 53.1/\underline{69.4} & 20.8/43.7 & 10.2/\textbf{18.9} & 32.0/50.2 \\
\ \  +PODS   & \textbf{12.2}/\underline{30.6} & 5.9/\underline{22.3} & \underline{47.4}/75.1 & 77.0/\underline{94.2} & \underline{53.2}/69.3 & \underline{21.8}/\underline{44.0} & \underline{10.3}/18.3 & \underline{32.5}/\underline{50.5} \\
\rowcolor{mylightblue}\ \  \textbf{+D$^3$S} & \underline{11.2}/\textbf{32.2} & \textbf{6.9/24.0} & \textbf{48.6/79.7} & \textbf{77.5}/94.1 & \textbf{53.7/69.5} & \textbf{23.5/44.5} & \textbf{10.6}/\underline{18.4} & \textbf{33.1/51.8} \\ \hline
GSPO    & 11.1/29.9 & 6.9/23.0 & \textbf{49.7}/\textbf{79.8} & \textbf{78.1}/\textbf{94.3} & \underline{53.5}/\underline{69.4} & \textbf{23.0/44.1} & \underline{10.5}/\textbf{19.4} & \underline{33.3}/\underline{51.4} \\
\ \  +PODS   & \textbf{12.3}/\textbf{32.9} & \underline{6.9}/\underline{24.0} & 47.8/77.1 & 77.4/\underline{94.1} & 53.4/\underline{69.4} & 22.5/\underline{43.3} & 10.2/18.7 & 32.9/\underline{51.4} \\
\rowcolor{mylightblue}\ \  \textbf{+D$^3$S} & 11.4/32.8 & \textbf{8.2/25.2} & \underline{48.4}/\underline{79.1} & \underline{78.0}/\underline{94.1} & \textbf{54.0/69.6} & \underline{22.9}/43.2 & \textbf{10.5}/\underline{19.0} & \textbf{33.3/51.9} \\ \hline
\multicolumn{9}{c}{Llama3.1-8B-Instruct} \\ \hline
Base &
  1.7/10.9 &
  \textbf{0.4/2.8} &
  \underline{15.0}/\underline{47.3} &
  57.7/92.8 &
  29.3/\underline{55.9} &
  14.7/\underline{38.5} &
  \underline{2.2}/\underline{8.2} &
  17.3/\underline{36.6} \\
GRPO &
  2.0/5.0 &
  0.0/0.0 &
  13.7/33.4 &
  \underline{78.6}/\underline{93.5} &
  \underline{31.5}/52.0 &
  15.9/35.6 &
  2.1/7.2 &
  20.5/32.4 \\
\ \  +PODS &
  \underline{2.8}/\underline{9.9} &
  \underline{0.3}/\underline{2.5} &
  14.5/38.1 &
  77.1/93.3 &
  \underline{31.5}/52.6 &
  \underline{16.3}/38.0 &
  \underline{2.2}/7.6 &
  \underline{20.7}/34.6 \\
\rowcolor{mylightblue}\ \  +\textbf{D$^3$S} &
  \textbf{5.3/20.7} &
  0.1/0.8 &
  \textbf{20.3/50.8} &
  \textbf{79.0/95.0} &
  \textbf{35.9/59.2} &
  \textbf{22.5/44.3} &
  \textbf{3.3/10.7} &
  \textbf{23.8/40.2} \\ \hline

\end{tabular}
}
\end{table}

Table~\ref{Table:Pass_1_8} presents the alignment results of the different LLMs across seven reasoning benchmarks, using GRPO and GSPO as the base algorithms. Our observations are four folds. \textbf{First}, the introduction of D$^3$S consistently improves performance across all backbone models, demonstrating its strong adaptability across various types of backbones. Notably, D$^3$S achieves the highest average scores of 35.8\% for pass@1 and 56.8\% for pass@8 on the Qwen2.5-Math-7B model. \textbf{Second}, D$^3$S demonstrates a significant performance advantage over both the original method and PODS. For example, when Qwen2.5-Math-7B is used as the backbone, GRPO+D$^3$S achieves an average improvement of 4.5 points in pass@1 and 3.7 points in pass@8 compared to the original GRPO. Similarly, with Llama3.1-8B-Instruct as the backbone, GRPO with D$^3$S surpasses the original GRPO and GRPO+PODS by 3.3 and 3.1 points on pass@1, and by 7.8 and 5.6 points on pass@8, respectively. \textbf{Third}, even for the strong baseline GSPO, incorporating D$^3$S still yields improvements of 2.6 and 1.5 points on pass@1 and pass@8, respectively. This demonstrates that the D$^3$S down-sampling strategy can be seamlessly generalized to and further enhance other algorithms leveraging group-relative advantages. \textbf{Fourth}, for the smaller-scale Qwen2.5-Math-1.5B model, D$^3$S still achieves the best average performance among all compared methods, showing its scalability and effectiveness across LLMs of varying sizes. We provide more experimental results, including pass@16 metrics, in Appendix~\ref{Section:Pass_16}, which further validate the effectiveness of our approach.

\subsection{Ablation Study of \texorpdfstring{D$^3$S}{D3S}}
\label{Section:Ablation}
D$^3$S integrates sample-level, token-level, and a dynamic down-sampling schedule to effectively train the policy model. To systematically evaluate the impact of each component in the D$^3$S strategy, we conduct a series of ablation studies by progressively incorporating more sophisticated selection strategies into the GRPO baseline, as listed in Table~\ref{Table:Ablation}. Here, we utilize Qwen2.5-Math-7B as the base model. The notation for each method is explained in detail in Section~\ref{Section:Experiment_Configuration}.

\begin{table}[t]
\centering
\caption{Ablation studies of different components in D$^3$S strategy. Performance is evaluated using Pass@8 across various benchmarks. The experiments utilize Qwen2.5-Math-7B as the base model and GRPO as the foundational algorithm.}
%\caption{Ablation study by incrementally applying different part of strategies of D$^3$S. Performance across benchmarks measured in pass@8. Experiments are implemented with Qwen2.5-Math-7B as base model and GRPO as base algorithm.}
\label{Table:Ablation}
\setlength{\tabcolsep}{3pt}
\setlength{\extrarowheight}{2pt}
\resizebox{0.8\textwidth}{!}{
\begin{tabular}{lcccccccc}
\hline
\textbf{Model}            & \textbf{AIME24} & \textbf{AIME25} & \textbf{AMC23} & \textbf{GSM8k} & \textbf{MATH} & \textbf{Minerva} & \textbf{Olympiad} & \textbf{Average} \\ \hline
% \multicolumn{9}{c}{Qwen2.5-Math-7B}                                                               \\ \hline
Base            & 8.9/33.2  & 2.3/13.4 & 22.8/70.4 & 30.1/83.2 & 27.9/64.6 & 8.4/33.7      & 4.1/14.6      & 14.9/44.7     \\
GRPO            & 13.2/37.6 & 5.5/21.6 & 47.0/83.5 & 64.9/94.3 & 48.5/70.2 & 19.8/45.0     & 9.7/19.8      & 29.8/53.1     \\
\ \  +D$^1$S          & 13.2/42.9 & 5.9/20.2 & 50.6/84.4 & 68.5/94.9 & 50.1/70.5 & 20.8/46.4     & 10.3/20.4     & 31.3/54.2     \\
\ \  +D$^1$S-C & 17.3/40.0 & 7.7/25.6 & 51.9/83.3 & \textbf{73.4}/95.4 & 52.8/70.9 & \textbf{25.0}/47.1     & 10.6/20.6     & 34.1/54.7     \\
\ \  +D$^2$S          & 16.9/42.2 & 6.0/21.2 & 49.6/82.8 & 66.3/94.9 & 49.5/70.7 & 20.9/46.7     & 10.1/20.3     & 31.3/54.1     \\
\rowcolor{mylightblue}\ \  +\textbf{D$^3$S} & \textbf{20.3/48.2} & \textbf{7.9/25.8} & \textbf{54.4/87.1} & 71.3/\textbf{95.7} & \textbf{52.2/71.5} & 23.4/\textbf{48.2} & \textbf{10.7/20.8} & \textbf{34.3/56.8} \\ \hline
\end{tabular}
}
\vspace{-3mm}
\end{table}

%D$^3$S contains a joint set of strategies to robustly guide model during training. In order to systematically investigate the effect of different part of D$^3$S strategy, a set of ablation study is implemented through increasingly add complicated selection strategy to GRPO base algorithm, as shown in Table~\ref{Table:Ablation}. Notations of methods have been detailed in Section~\ref{Section:Experiment_Configuration}. 

The ablation results reveal that the contribution of individual components is not strictly monotonic. For instance, D$^2$S occasionally underperforms D$^1$S on AIME24 but achieves greater improvements on AIME25. Nevertheless, the complete D$^3$S configuration consistently delivers the best performance. This provides strong evidence that the dual-level design, combined with a dynamic down-sampling schedule, effectively balances exploitation and exploration, leading to robust improvements across tasks. Beyond the Qwen2.5-Math-7B model, additional ablation studies and analyses on Llama3.1 models, detailed in Section~\ref{Section:Ablation_Llama}, further demonstrate the generalizability of the D$^3$S.

%The ablation study of D$^3$S shows that adding each strategy component is not strictly monotonic, e.g. D$^2$S sometimes slightly lags behind D$^1$S on AIME24 while gaining more improvement on AIME25. But the full D$^3$S configuration consistently delivers the best results. This result is a strong evidence suggesting that the dual-level design with a dynamic schedule effectively balances exploitation and exploration, providing robust improvements across tasks. In addition to Qwen2.5-Math-7B model, ablation study and analysis about Llama3.1 models are detailed in Section~\ref{Section:Ablation_Llama} as extensive evidence of generality of D$^3$S framework. Figure~\ref{Figure:Ablation_Llama} strongly suggests the potential of D$^3$S in improving whole optimization procedure through large-enough but smoother policy gradient and lower $\KL$ cost.

\begin{table}[ht]
\centering
\caption{Comparison of training efficiency. We assess the performance gains brought by integrating D$^3$S into the GRPO and GSPO, along with the time acceleration needed to achieve these gains.
%Training efficiency measured in maximum Avg@32 metric and time consumed to achicheve such performance
}
\label{Table:Training_Efficiency}
\resizebox{0.6\textwidth}{!}{%
\begin{tabular}{lcccc}
\hline
\multicolumn{1}{c}{\multirow{2}{*}{\textbf{Methods}}} & \multicolumn{2}{c}{\textbf{D$^3$S vs GRPO}} & \multicolumn{2}{c}{\textbf{D$^3$S vs GSPO}} \\
\multicolumn{1}{c}{} & Avg@32 & Time         & Avg@32 & Time         \\ \hline
Qwen2.5-Math-7B      & \textbf{+6\%}   & \textbf{2.04$\times$} & \textbf{+17\%}  & \textbf{5.51$\times$} \\
Qwen2.5-Math-1.5B    & +4\%   & 1.57$\times$ & +2\%   & 1.10$\times$ \\ \hline
\end{tabular}%
}
\end{table}

\vspace{-4mm}
\subsection{Training efficiency}

%Table~\ref{Table:Training_Efficiency} shows the significant advantages of D$^3$S in training efficiency. Taking Qwen2.5-Math-7B as an example, while D$^3$S achieves an average accuracy (Avg@32) that is 6\% higher than GRPO, the time required to achieve the same performance as GRPO is only half that of GRPO (2.04$\times$ acceleration). Compared with GSPO, the improvement is even more significant, with a 5.51$\times$ increase in training speed and a 17\% increase in accuracy. On the smaller Qwen2.5-Math-1.5B model, D$^3$S also brings a 4\% performance improvement and a 1.57$\times$ acceleration effect. Overall, D$^3$S significantly accelerates the convergence of the model, while further improving the final performance, fully demonstrating its universality and practical value on models of different scales.
Table~\ref{Table:Training_Efficiency} highlights the remarkable training efficiency of D$^3$S. For instance, on the Qwen2.5-Math-7B model, D$^3$S achieves an average accuracy (Avg@32) that is 6\% higher than GRPO, while reducing the training time by half to reach the same performance level (2.04$\times$ speedup). The benefits are even more pronounced compared to GSPO, with a 5.51$\times$ faster training speed and a 17\% accuracy improvement. On the smaller Qwen2.5-Math-1.5B model, D$^3$S delivers a 4\% accuracy boost alongside a 1.57$\times$ speedup. These findings underscore D$^3$S's ability to not only accelerate model convergence but also enhance final performance, particularly for larger models.

\subsection{Dynamic down-sampling schedule mitigates overfitting}

\vspace{-2mm}
\begin{wrapfigure}{r}{0.4\textwidth}
    \centering
    \vspace{-15pt}
    \begin{subfigure}{\linewidth}
        \includegraphics[width=\linewidth]{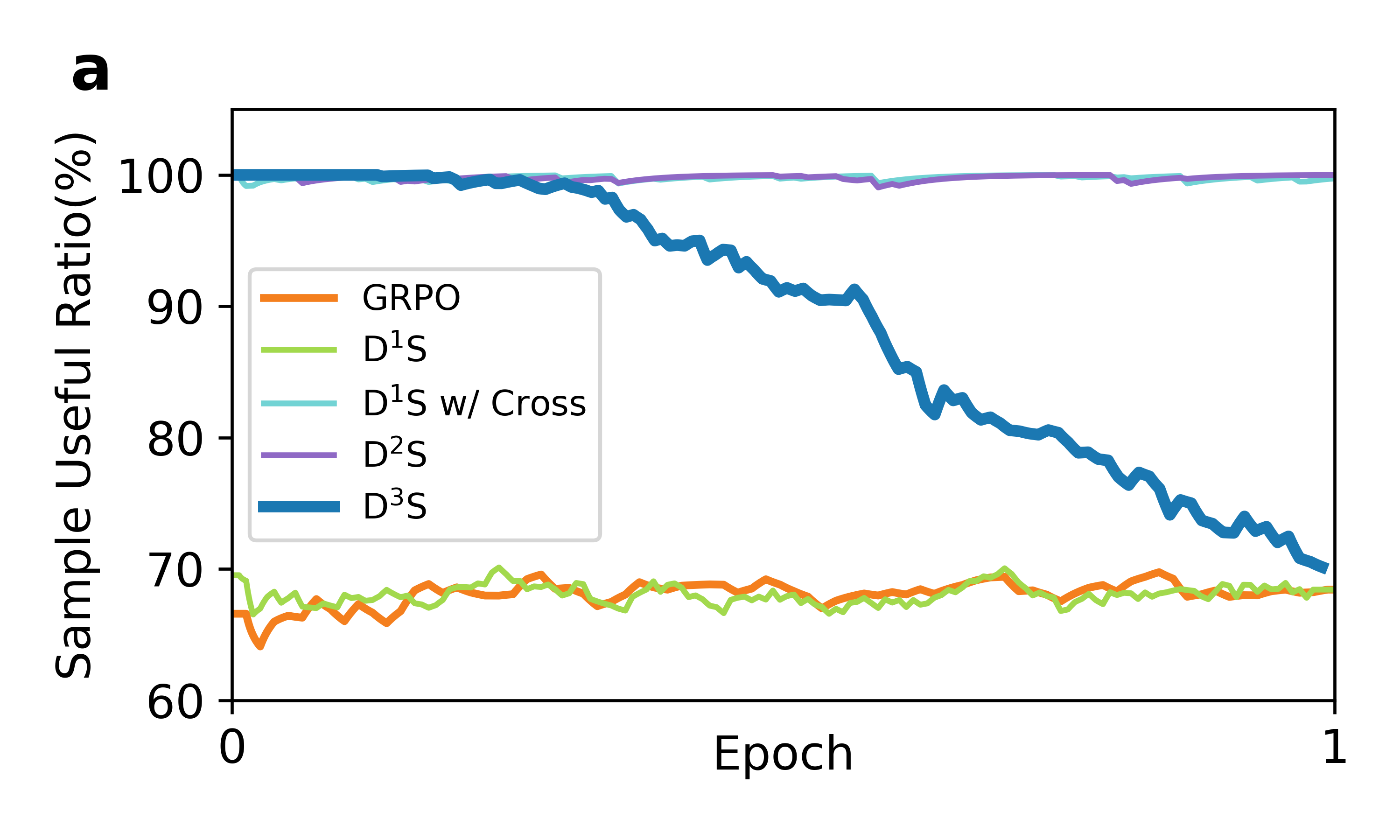}
        \caption{}
        \label{Figure:Train_Sample_Useful_Train_Qwen_ema}
    \end{subfigure}

    \vspace{-10pt}
    \begin{subfigure}{\linewidth}
        \includegraphics[width=\linewidth]{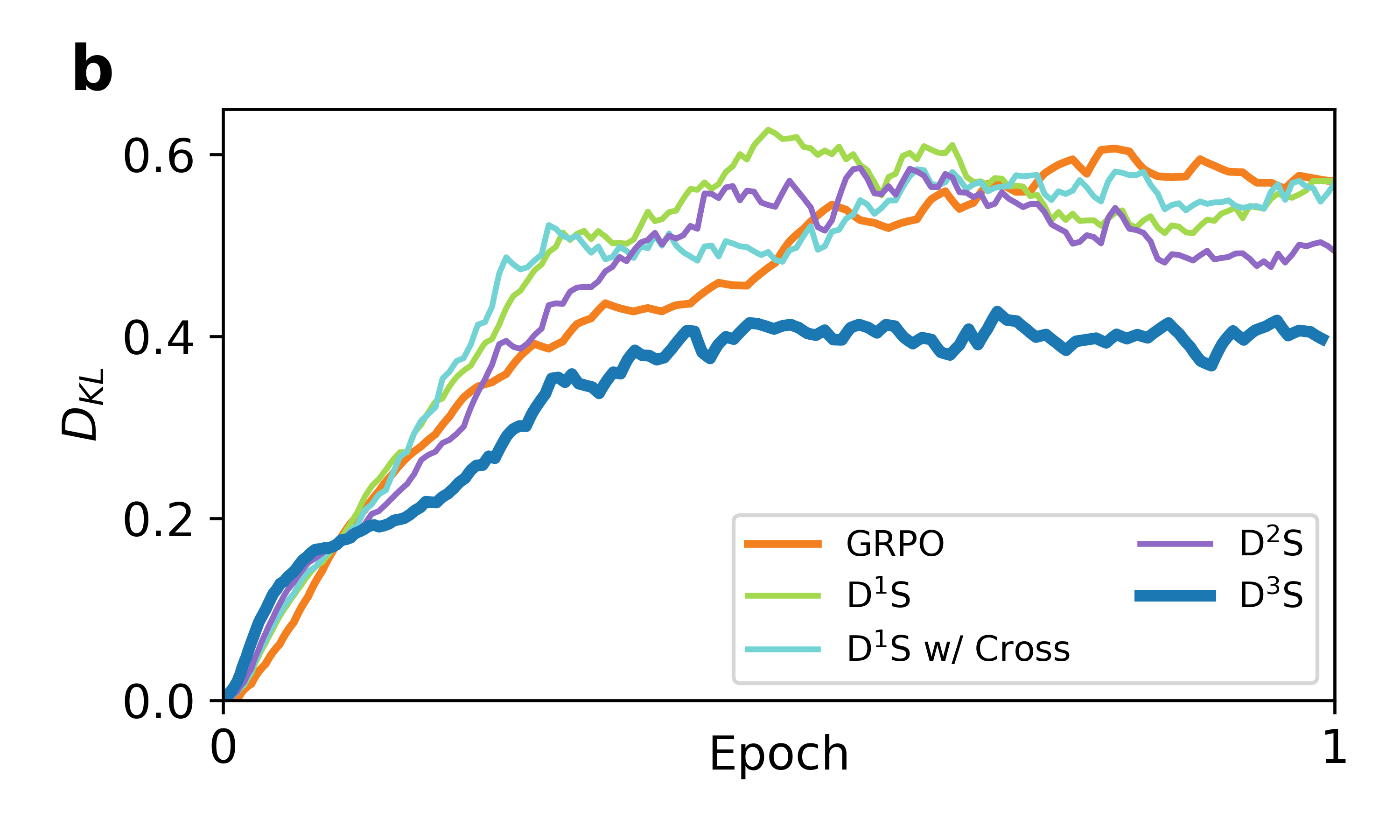}
        \caption{}
        \label{Figure:Train_KL_Qwen_ema}
    \end{subfigure}
    \vspace{-9mm}
    \caption{Training dynamics of (\subref{Figure:Train_Sample_Useful_Train_Qwen_ema}) sample usefulness and (\subref{Figure:Train_KL_Qwen_ema}) KL divergence. 
    % SUR measures the proportion of samples in a batch that actively contribute to policy updates. $\KL$quantifies the deviation of the policy from the reference model distribution.
    }
    %\caption{Useful samples (samples of which advantages are non-zero) ratio in training (\subref{Figure:Train_Sample_Useful_Train_Qwen_ema}) and $\KL$ metrics (\subref{Figure:Train_KL_Qwen_ema}). Once cross group down sampling is introduced, model update parameters focusing on samples with non-zero advantage signals. D$^3$S dynamically schedule learning criteria so as to learn more diverse data without sacrificing $\KL$.}
    \label{Figure:Sample_useful_and_KL}
    \vspace{-20pt}
\end{wrapfigure}
To better understand how D$^3$S impacts policy optimization, we track key metrics during the training process. The first metric, sample usefulness rate (SUR), measures the proportion of groups in each batch with non-zero advantages, reflecting the percentage of samples that actively contribute to policy updates. The second metric, \textit{KL divergence} ($\KL$), quantifies the difference between the policy and reference model distributions. A higher $\KL$ indicates greater divergence from the reference model, which may signal an increased risk of overfitting.

Figure~\ref{Figure:Sample_useful_and_KL} illustrates the training dynamics under various D$^3$S settings, as detailed in Section~\ref{Section:Experiment_Configuration}. Four key observations emerge: \textbf{First}, as shown in Figure~\ref{Figure:Train_Sample_Useful_Train_Qwen_ema}, both the original GRPO and D$^1$S (without cross-group operation) maintain a SUR of approximately 70\%, with minor fluctuations. \textbf{Second}, introducing cross-group operation (e.g., D$^1$S w/Cross and D$^2$S) significantly boosts the SUR to nearly 100\%, indicating that cross-group mechanism effectively filter out ambiguous data within the training batch. \textbf{Third}, as shown in Figure~\ref{Figure:Train_KL_Qwen_ema}, the $\KL$ curves for D$^1$S, D$^1$S w/Cross, and D$^2$S rise more steeply in the early stages compared to GRPO but eventually converge to similar values. This suggests that while these methods initially deviate more from the reference model, they ultimately reach comparable limits, potentially increasing the risk of overfitting. \textbf{Fourth}, the SUR gradually declines from 100\% to 70\% as training progresses, aligning with the design goal of dynamic down-sampling—incrementally increasing samples and tokens to mitigate overfitting in later stages. The $\KL$ curve also demonstrates that D$^3$S achieves smaller deviations from the reference model compared to other methods.  

\begin{figure}[ht]
    \centering
    \begin{subfigure}{0.24\textwidth}
        % 设定图片宽度为文本宽度的0.24倍，留出间隙
        \includegraphics[width=\linewidth]{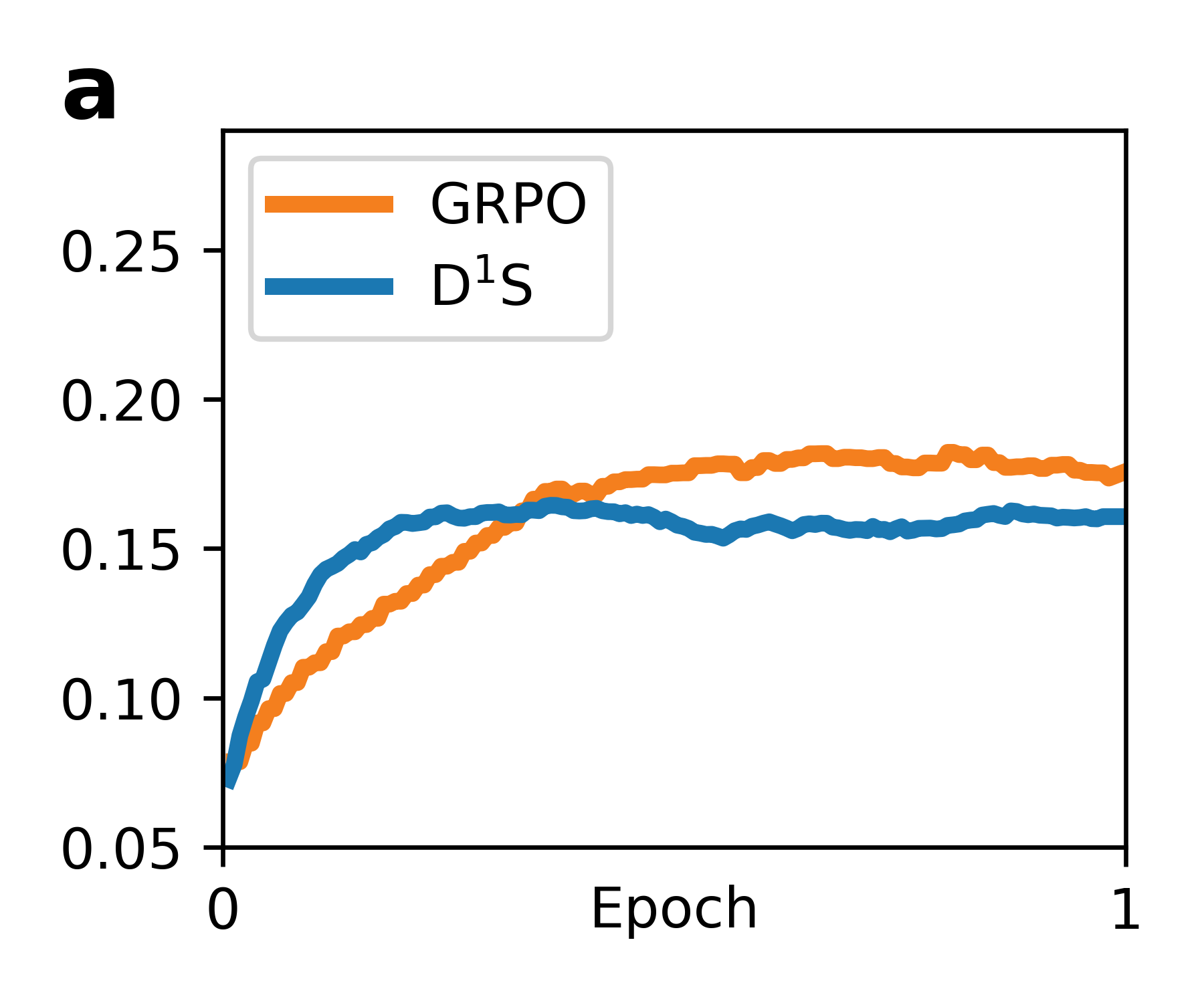}
        \caption{}
        \label{Figure:Train_Qwen_1D-inner} % 为子图(a)添加标签，方便引用
    \end{subfigure}
    % \hfill % 这个命令是关键，它会在子图之间创建弹性的水平间距
    % 第二个子图
    \begin{subfigure}{0.24\textwidth}
        \includegraphics[width=\linewidth]{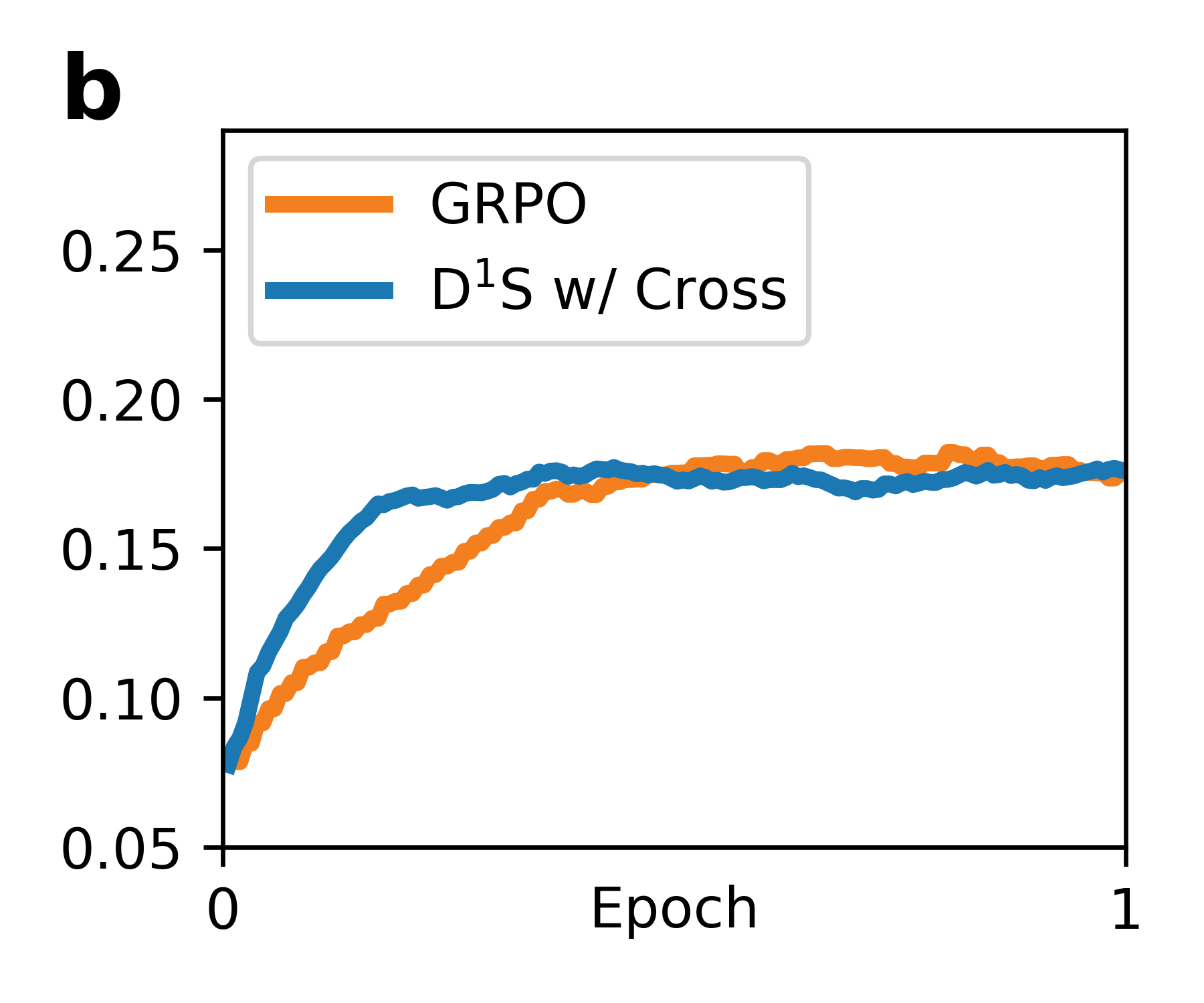}
        \caption{}
        \label{Figure:Train_Qwen_1D-Cross}
    \end{subfigure}
    % \hfill % 同样，在第二个和第三个子图之间添加间距
    % 第三个子图
    \begin{subfigure}{0.24\textwidth}
        \includegraphics[width=\linewidth]{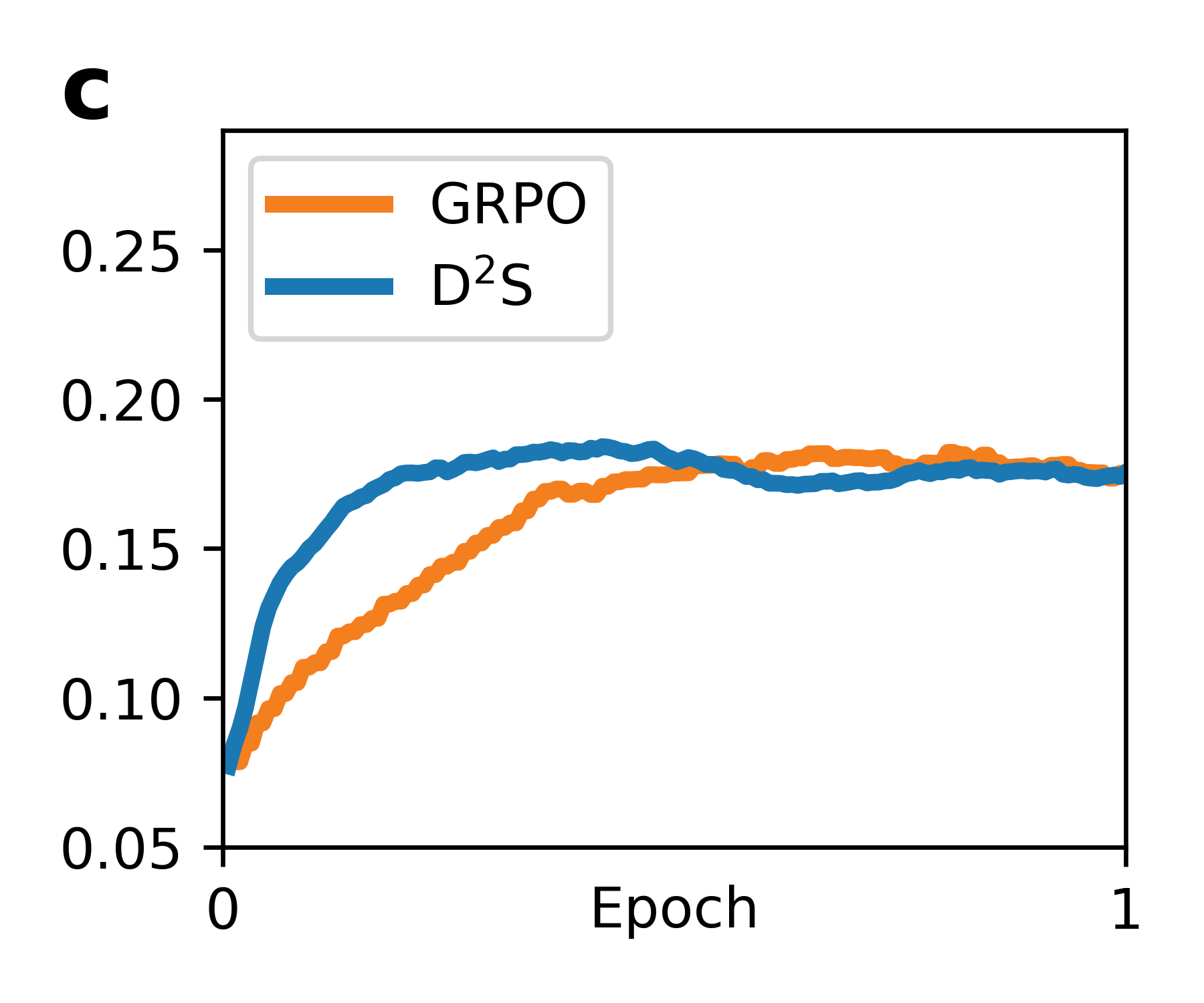}
        \caption{}
        \label{Figure:Train_Qwen_2D}
    \end{subfigure}
    % \hfill % 在第三个和第四个子图之间添加间距
    % 第四个子图 (注意：最后一个子图后面不需要 \hfill)
    \begin{subfigure}{0.24\textwidth}
        \includegraphics[width=\linewidth]{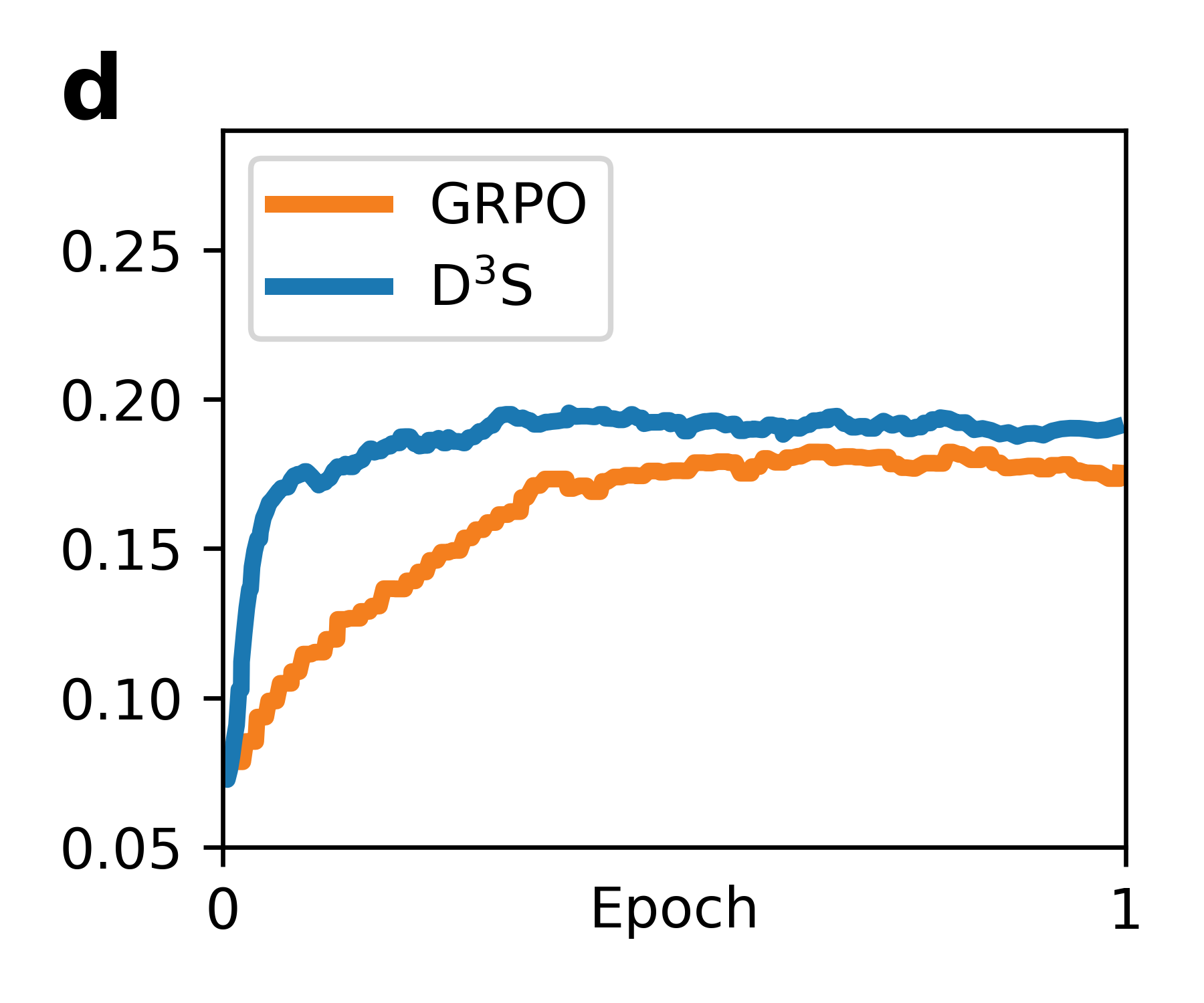}
        \caption{}
        \label{Figure:Train_Qwen_D3S}
    \end{subfigure}
    \vspace{-2mm}
    \caption{The Avg@32 test performance of AIME24 on Qwen2.5-Math-7B under various settings. Methods without a dynamic down-sampling schedule (\subref{Figure:Train_Qwen_1D-inner},\subref{Figure:Train_Qwen_1D-Cross},\subref{Figure:Train_Qwen_2D}) accelerate convergence in the early stages but suffer from overfitting later. In contrast, the dynamic down-sampling schedule (\subref{Figure:Train_Qwen_D3S}) not only accelerates convergence initially but also outperforms other methods in the later stages. }
    %\caption{Static aggressive selection strategies (\subref{Figure:Train_Qwen_1D-inner},\subref{Figure:Train_Qwen_1D-Cross},\subref{Figure:Train_Qwen_2D}) suffer from overfitting, while dynamic down-sampling (\subref{Figure:Train_Qwen_D3S}) consistently stabilizes training on Qwen2.5-Math-7B. This in-procedure training metric Avg@32 is collected from average accuracy of 32 validating runs on AIME24 task during training.}
        \label{Figure:Dynamic_vs_Static_Qwen}
\end{figure}
\vspace{2mm}

Figure~\ref{Figure:Dynamic_vs_Static_Qwen} provides a further comparison of Avg@32 performance under different down-sampling configurations. It can be observed that down-sampling strategies without a dynamic schedule consistently accelerate convergence but exhibit varying degrees of overfitting, with their Avg@32 accuracies eventually being surpassed by GRPO in the later stages of training (Figure~\ref{Figure:Train_Qwen_1D-inner},~\subref{Figure:Train_Qwen_1D-Cross},~\subref{Figure:Train_Qwen_2D}). In contrast, D$^3$S (Figure~\ref{Figure:Train_Qwen_D3S}), which incorporates a dynamic schedule, not only accelerates convergence in the early stages but also achieves significantly better results in the later stages. This highlights the critical role of dynamic down-sampling schedule in mitigating overfitting.

\subsection{Entropy Analysis of \texorpdfstring{D$^3$S}{D3S}}
\begin{figure}[b]
    \centering
    \begin{subfigure}{0.24\textwidth}
        \includegraphics[width=\linewidth]{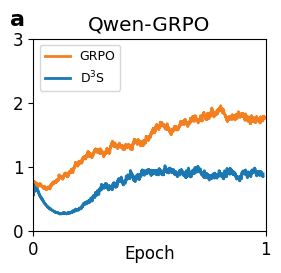}
        \caption{}
        \label{Figure:Entropy_Qwen-GRPO}
    \end{subfigure}
    % \hfill 
    \begin{subfigure}{0.24\textwidth}
        \includegraphics[width=\linewidth]{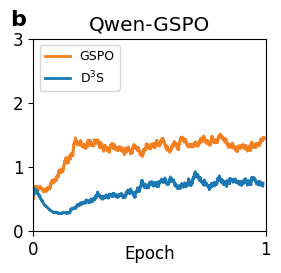}
        \caption{}
        \label{Figure:Entropy_Qwen-GSPO}
    \end{subfigure}
    % \hfill
    \begin{subfigure}{0.24\textwidth}
        \includegraphics[width=\linewidth]{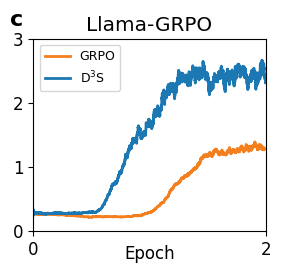}
        \caption{}
        \label{Figure:Entropy_Llama-GRPO}
    \end{subfigure}
    % \hfill
    \begin{subfigure}{0.24\textwidth}
        \includegraphics[width=\linewidth]{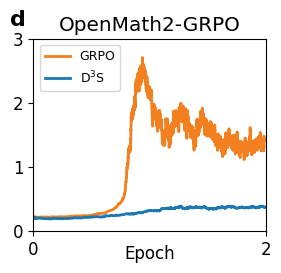}
        \caption{}
        \label{Figure:Entropy_OpenMath2-GRPO}
    \end{subfigure}
    \caption{Entropy dynamics of different base models and RL algorithms. D$^3$S effectively balances exploration and exploitation, fostering confident policies in well-aligned backbones (\subref{Figure:Entropy_Qwen-GRPO}, \subref{Figure:Entropy_Qwen-GSPO}, and \subref{Figure:Entropy_OpenMath2-GRPO}) while driving essential exploration in less-aligned ones (\subref{Figure:Entropy_Llama-GRPO}).}
        \label{Figure:Scaled_entropy}
\end{figure}

We investigate entropy dynamics across various base models and RL algorithms to understand learning behaviors. As illustrated in Figures~\ref{Figure:Entropy_Qwen-GRPO} and \ref{Figure:Entropy_Qwen-GSPO}, D$^3$S consistently achieves lower and more stable policy entropy compared to baseline algorithms. This improvement stems from the token-level selection mechanism described in Section~\ref{Section:Token_level}. By prioritizing updates on tokens with a high advantage-entropy product ($|A_{i,t}|\times H_{i,t}$), D$^3$S focuses learning on resolving high-impact uncertainties. This targeted strategy enables the model to converge more efficiently toward a decisive policy, as evidenced by its reduced average entropy.

Conversely, the Llama3.1-8B-Instruct model exhibits a distinct behavior, as illustrated in Figure~\ref{Figure:Entropy_Llama-GRPO}. We hypothesize that this phenomenon arises from the base capabilities of Llama3.1 and its lack of adaptation to mathematical reasoning tasks, which leads D$^3$S to promote more effective and productive exploration. To validate this hypothesis, we introduce OpenMath2-Llama3.1-8B \citep{toshniwal2024openmath2}, a model fine-tuned specifically for mathematical tasks, as a point of comparison. As illustrated in Figure~\ref{Figure:Entropy_OpenMath2-GRPO}, D$^3$S exhibits entropy dynamics consistent with those observed in Figures~\ref{Figure:Entropy_Qwen-GRPO} and \ref{Figure:Entropy_Qwen-GSPO}, reinforcing our hypothesis. While the baseline GRPO shows a sharp and unstable entropy spike early in training, D$^3$S ensures a smoother and more controlled learning trajectory with consistently low entropy. Additional analyses are provided in Appendix~\ref{Section:Ablation_Llama}.

In conclusion, the entropy dynamics reveal that D$^3$S effectively balances exploration and exploitation, fostering confident policies in aligned models while encouraging necessary exploration in less-aligned ones. This further underscores the robustness of the proposed framework.

\section{Related Works}

\paragraph{Data Selection for Enhancing Training Efficiency}  
As models and datasets scale, selecting data with high informational value becomes crucial for improving training efficiency. \citet{razin2024VanishingGradientsReinforcementa,razin2025WhatMakesReward} suggests that increasing reward signal variance sharpens the optimization landscape, thereby accelerating convergence. \citet{lu2024ItTakesTwo} introduces SEAM, an automated metric for quantifying PM-RM differences, which enhances training reliability by filtering out samples where RM misjudges PM outputs. \citet{gou2025MixedPreferenceOptimization} proposes Mixed Preference Optimization (MPO), which first uses DPO training on "easy" data with large reward gaps, and then performs RLHF on "hard" data with small reward gaps. \citet{pattnaik2024CurryDPOEnhancingAlignment} presents Curry-DPO, a curriculum-based approach that organizes preference pairs by difficulty, transitioning from pairs with large response gaps to those with smaller differences. \citet{chen2025DataCentricSampleCentricEnhancing} proposes LearningProgress and Prefix-guided Optimization (LPPO), which dynamically adjusts sample weighting based on the model's learning progress. Their method emphasizes marginal samples nearing mastery while downweighting those already learned or excessively challenging. Unlike these approaches, our sample-level selection strategy maximizes advantage variance. we theoretically prove that it yields a higher gradient upper bound, thereby enhancing training efficiency.

\paragraph{Token-Level Entropy Utilization}
\citet{cui2025EntropyMechanismReinforcement} empirically establishes a link between performance improvement, policy entropy reduction, and exploration capacity consumption. Entropy reflects the unequal importance of tokens within a sequence. \citet{wang20258020Rule} observes that high-entropy tokens disproportionately influence the policy gradient. Furthermore, \citet{cui2025EntropyMechanismReinforcement} shows that changes in policy entropy are driven by the covariance between action probabilities and advantage values. To address this, they propose Clip-Cov and KL-Cov, which mitigate entropy loss by clipping or penalizing updates to tokens with high positive covariance. Building on this, \citet{wen2024EntropyRegularizedTokenLevelPolicy} introduces Entropy-Regularized Token-Level Policy Optimization (ETPO) to improve credit assignment. Similarly, \citet{shen2025EntropyControlLLMRL} proposes AEnt, which calculates clamped entropy over a subset of high-probability tokens and employs an adaptive coefficient to balance the entropy reward. These approaches can be interpreted as intelligent budget allocation, focusing entropy reduction on the most impactful regions of the policy space. In our study, we integrate entropy with the magnitude of advantage, prioritizing optimization on tokens that are both critical and highly uncertain.

\section{Conclusion}
In this study, we reveal that the theoretical upper bound of the policy gradient norm in group-relative advantage-based algorithms (e.g., GRPO) is positively correlated with advantage variance. Leveraging this insight, we introduce the Dynamic Dual-level Down-Sampling (D$^3$S) framework to enhance training efficiency. At the sample-level, D$^3$S selects rollout responses to maximize advantage variance, while at the token-level, it prioritizes tokens with a high product of entropy and advantage magnitude, directing updates to regions where the model is both uncertain and impactful. To mitigate overfitting, D$^3$S adopts a dynamic down-sampling schedule inspired by curriculum learning, gradually relaxing sampling criteria over time. Extensive experiments on the Qwen2.5 and Llama3.1 models show that D$^3$S consistently surpasses baseline methods across diverse reasoning benchmarks. Ablation studies and dynamic analyses reveal that the synergy between its components is crucial for balancing exploration and exploitation. These results highlight the significance of fine-grained sample utilization in RLHF and emphasize the pivotal role of entropy in managing the exploration-exploitation trade-off.

\section*{Ethics Statement}  
% \vspace{-0.2cm}
Our study introduces the Dynamic Dual-Level Down-Sampling (D$^3$S) framework to improve the efficiency and performance of RL algorithms. We affirm that this research raises no significant ethical concerns. All methodologies strictly adhere to ethical standards and responsible research practices. The datasets used are publicly available, widely recognized within the research community, and utilized in full compliance with their terms and conditions. Additionally, we declare no conflicts of interest, sponsorships, or external influences that could compromise the integrity of this work. In line with our commitment to transparency and reproducibility, we will release our code publicly to facilitate further research and innovation in this domain.

\section*{Reproducibility Statement}
% \vspace{-0.2cm}

Experimental details are presented in Section \ref{Section:Experiment_Configuration}. The code and data used in this study are included in the supplementary materials and will be publicly released. Proofs for the main theoretical results (Propositions \ref{Proposition:J_GRPO}, \ref{Proposition:J_GRPO_subset}, and Lemma \ref{Lemma:Max_Var}) can be found in Appendices \ref{Section:Proof_Proposition_1}, \ref{Section:Proof_Proposition_2}, and \ref{Section:Proof_Lemma_1}, respectively.

%In this work, we revealed the theoretical upper bound of policy gradient norm in group-relative advantage-based algorithms (e.g. GRPO) is positively correlated with variance of advantages. To accelerate policy optimization, we introduced the Dynamic Dual-level Down-Sampling ($D^{3}S$) framework to enhance exploitation of most-valuable learning signals while maintaining exploration potential. On the sample-level, $D^{3}S$ selects rollout responses to maximize advantage variance ($\text{Var}(A)$),. On the token-level, $D^{3}S$ refines exploration by prioritizing tokens with a high advantage-entropy product ($|A_{i,t}|\times H_{i,t}$), directing updates to where the model is most uncertain yet the outcome is most impactful. To mitigate the risk of overfitting, $D^{3}S$ incorporates a dynamic schedule inspired by curriculum learning that gradually relaxes sampling criteria,. Our comprehensive experiments on Qwen2.5 and Llama3.1 models demonstrate that $D^{3}S$ consistently outperforms baseline group-relative advantage-based methods across multiple reasoning benchmarks. Ablation studies and dynamic analyses confirmed that the synergy between its components is crucial for effectively managing the exploration-exploitation trade-off. Our findings highlight the potential of fine-grained sample utilization in RLHF and the significancy of entropy in balancing exploitation and exploration.
% \newpage

\bibliography{iclr2026_conference}
\bibliographystyle{iclr2026_conference}

\newpage
\appendix

\section{Usage of Large Language Models}

In this study, we leverage LLMs to summarize and refine academic papers, ensuring clarity, precision, grammatical accuracy, and correct spelling. These models also offer suggestions to improve coherence and readability. Our aim is to elevate the efficiency and quality of academic writing.

\section{Proof}\label{Section:Proof}
\subsection{Proof of Proposition~\ref{Proposition:J_GRPO}}\label{Section:Proof_Proposition_1}
\begin{proof}[Proof of Proposition~\ref{Proposition:J_GRPO}]
We analyze the gradient at the reference policy without considering the clipping operation, as it is inactive when the importance ratios equal 1.

The gradient becomes:
\begin{align}
\nabla_\vtheta J_{\text{GRPO}}(\vtheta) 
&= \E_{x \sim \sD, {y_i}_{i=1}^G \sim \pi_{\vtheta}(\cdot|x)} \left[ \frac{1}{G} \sum_{i=1}^G \frac{1}{|y_i|} \sum_{t=1}^{|y_i|} A_{i,t} \nabla_\vtheta \log \pi_\vtheta(y_{i,t}|x, y_{i,<t}) \right] \\
&= \E_{x \sim \sD, {y_i}_{i=1}^G \sim \pi_{\vtheta}(\cdot|x)} \left[ \frac{1}{G} \sum_{i=1}^G \frac{1}{|y_i|} \sum_{t=1}^{|y_i|} A_{i,t} \nabla_\vtheta \log \text{softmax}(z_\vtheta(y_{i,t}|x, y_{i,<t})) \right] \\
&= \E_{x \sim \sD, {y_i}_{i=1}^G \sim \pi_{\vtheta}(\cdot|x)}& \nonumber \\
&  \left[ \frac{1}{G} \sum_{i=1}^G \frac{1}{|y_i|} \sum_{t=1}^{|y_i|} A_{i,t} (\mathrm{I}_{i,t}^{\text{one-hot}}-\pi_\vtheta(\cdot|x, y_{i,<t}))^T \nabla_\vtheta z_\vtheta(y_{i,t}|x, y_{i,<t})  \right] 
\end{align}

In following analysis, we simplify the average of token-wise advantages $\sum_{t=1}^{|y_i|} A_{i,t}$ into sequence-wise advantage $A_i$. On the other hand, it is also a description of the more commonly used sequence-wise outcome-rewarded GRPO implementations in math tasks. Formally:
\begin{align}
\nabla_\vtheta J_{\text{GRPO}}(\vtheta) = \E_{x \sim \sD, {y_i}_{i=1}^G \sim \pi_{\vtheta}(\cdot|x)} \left[
\frac{1}{G} \sum_{i=1}^G A_i \nabla_\vtheta \log \pi_\vtheta(y_i|x)
\right]
\end{align}

where $\nabla_\vtheta \log \pi_\vtheta(y_i|x) = \frac{1}{|y_i|} \sum_{t=1}^{|y_i|} \nabla_\vtheta \log \pi_\vtheta(y_{i,t}|x, y_{i,<t})$.

For any $c > 0$, we decompose the sum based on the magnitude of the standardized advantage:
\begin{align}
\nabla_\vtheta J_{\text{GRPO}}(\vtheta) &= \E \left[
\frac{1}{G} \sum_{i: |A_i| \leq c}^G A_i \nabla_\vtheta \log \pi_\vtheta(y_i|x)
\right] \tag{I} \label{Eqaution:Part_1_definition}\\
& + \E \left[
\frac{1}{G} \sum_{i: |A_i| > c}^G A_i \nabla_\vtheta \log \pi_\vtheta(y_i|x)
\right] \tag{II} \label{Eqaution:Part_2_definition}
\end{align}
Bounding term \ref{Eqaution:Part_1_definition}: For samples with $|A_i| \leq c$:
\begin{align}
\|\text{(I)}\| &\leq \E \left[
\frac{1}{G} \sum_{i: |A_i| \leq c}^G |A_i| \cdot \|\nabla_\vtheta \log \pi_\vtheta(y_i|x)\|
\right] \\
&\leq \E \left[\frac{1}{G} \sum_{i: |A_i| \leq c}^G c \cdot\|(\mathrm{I}_{i}^{\text{one-hot}}-\pi_\vtheta(\cdot|x))^T \nabla_\vtheta z_\vtheta(y_{i}|x)\| \right] \\
&\leq \E \left[\frac{1}{G} \sum_{i: |A_i| \leq c}^G c \cdot\|(\mathrm{I}_{i}^{\text{one-hot}}-\pi_\vtheta(\cdot|x))^T\| \cdot \| \nabla_\vtheta z_\vtheta(y_{i}|x)\| \right] 
\label{Equation:Part_1}
\end{align}

Considering non-negative property of one-hot vector and likelihood vector, we can get:

\begin{align}
    \|\mathrm{I}_{i}^{\text{one-hot}}-\pi_\vtheta(\cdot|x)\|
    &\le \|\mathrm{I}_{i}^{\text{one-hot}}-\pi_\vtheta(\cdot|x)\|_{1} \le 2
    \label{Equation:One_Hot}
\end{align}

where $\|\cdot\|_{1}$ denotes the L1 norm.

Apply Equation~\ref{Equation:One_Hot} to Equation~\ref{Equation:Part_1}, we can get:

\begin{align}
\|\text{(I)}\| &\leq c \cdot 2 \cdot \gamma(x;\vtheta)
\end{align}

where $\gamma(x;\vtheta)$ denotes a static parameter related to input query $x$ and model parameters $\vtheta$.

Similarly, we apply Equation~\ref{Equation:One_Hot} and $\gamma(x;\vtheta)$ in bounding term \ref{Eqaution:Part_2_definition}:

\begin{align}
    \|\text{(II)}\|
    &= \E \left[ \frac{1}{G} \sum_{i: |A_i| > c}^G A_i \nabla_\vtheta \log \pi_\vtheta(y_i|x) \right] \\
    &\leq \E \left[ \frac{1}{G} \sum_{i: |A_i| > c} |A_i| |\nabla_\vtheta \log \pi_\vtheta(y_i|x)| \right] \\
    &\leq 2 \cdot \gamma \cdot \E \left[ \frac{1}{G} \sum_{i: |A_i| > c}^G |A_i| \right] \\
    &= 2 \cdot \gamma \cdot \E[|A_i| \cdot \mathbf{1}_{|A_i| > c}] \label{Eqaution:Part_2}
\end{align}

For samples with $|A_i| > c$, we use the refined Chebyshev inequality. Since $A_i$ is group-normalized with $\E[A_i] = 0$ and $\text{Var}(A_i) = 1$, we have $\E[A_i^2] = 1$. By the refined Chebyshev bound:
\begin{align}
\E[|A_i| \cdot \mathbf{1}_{|A_i| > c}] \leq \frac{\E[A_i^2]}{c} = \frac{1}{c}
\label{Equation:Chebyshev}
\end{align}

Therefore:
\begin{align}
\|\text{(II)}\| 
&\leq 2 \cdot \gamma(x;\vtheta) \cdot \frac{1}{c}
\end{align}

Combining the bounds:
\begin{align}
\|\nabla_\vtheta J_{\text{GRPO}}(\vtheta)\| &\leq \|\text{(I)}\| + \|\text{(II)}\| \\
&\leq c \cdot 2 \cdot \gamma(x;\vtheta) + \frac{2 \cdot \gamma(x;\vtheta)}{c} \\
&= 2\gamma(x;\vtheta) \left(c + \frac{1}{c}\right)
\end{align}

The right-hand side objective is minimized when $c = 1$, we can obtain:
\begin{align}
\|\nabla_\vtheta J_{\text{GRPO}}(\vtheta)\| \leq 2 \cdot\gamma(x;\vtheta) \cdot 2 = 4\gamma(x;\vtheta)
\end{align}

\end{proof}

\vspace{-10mm}
\subsection{Proof of Proposition~\ref{Proposition:J_GRPO_subset}}\label{Section:Proof_Proposition_2}

\begin{proof}[Proof of Proposition~\ref{Proposition:J_GRPO_subset}]

Considering Equation~\ref{Equation:Chebyshev}, the premise for it to be valid is $\E[A]=0, \text{Var}(A)=1$ thus $\E[A^2]=\text{Var}(A) + \E[A]^2 = 1$, which is property of standardized advantages. Thus, applying $\E[A^2]=1$ and $|A_i| >c$, we can proof Equation~\ref{Equation:Chebyshev} as: 

\begin{align}
    \E[|A_i| \cdot \mathbf{1}_{|A_i| > c}] 
    &= \int_{|A_i| >c} |A_i|f(A)dA \\
    &\le \int_{|A_i| >c} |A_i| \cdot \frac{|A_i|}{c} \cdot f(A)dA \\
    &= \frac{1}{c} \int_{|A_i| >c} |A_i|^2f(A)dA \\
    &\le \frac{1}{c} \int_{-\infty}^{\infty} |A_i|^2f(A)dA \\
    &= \frac{\E[A_i^2]}{c} = \frac{1}{c}
\end{align}

However, considering $|A|$-driven down-sampling, $\E[{A'}^2]=1$ no longer holds for subset $A'$. 

We first estimate the upper bound of $|A_i|$. In GRPO, advantages are standardized as Equation~\ref{Equation:Advantage}. When reward signals are fixed in set ${0,1}$, the maximum of $|A_i|$ within a group of size $G$ can be obtained when only $1$ sample $i$ is rewarded with $1/0$ with other $G-1$ samples rewarded with $0/1$ correspondingly. Formally:

\begin{align}
    |A_i|_{\text{max}} 
    &= \frac{R_{i} - \E[R]}{\text{std}[R]} = \frac{1 - \frac{1}{G}}{\sqrt{\E[R^2]-\E[R]^2}} \\
    &= \frac{1-\frac{1}{G}}{\frac{\sqrt{G-1}}{G}} = \sqrt{G-1} \label{Equation:Max_Advantage}
\end{align}

Apply Equation~\ref{Equation:Max_Advantage} to Equation~\ref{Eqaution:Part_2}, we can get:

\begin{align}
    \E[|A'_i| \cdot \mathbf{1}_{|A'_i| > c}] 
    &\le \sqrt{G - 1} \cdot \mathcal{P}(|A'_i| > c) \\
    \|\text{(II)}\| 
    &\leq 2 \cdot \gamma(x;\vtheta) \cdot \sqrt{G - 1} \cdot \frac{\text{Var}[A']}{c^2} \\
    \|\nabla_\vtheta J_{\text{GRPO}}(\vtheta)\| 
    &\leq 2 \cdot \gamma(x;\vtheta) \left(c + \sqrt{G - 1} \cdot \frac{\text{Var}(A')}{c^2}\right)
\end{align}

where we try to choose optimal $c$:

\begin{align}
\frac{d}{dc}\left(c + \sqrt{G - 1} \cdot \frac{\text{Var}(A')}{c^2}\right) 
&= 1 - 2\sqrt{G - 1} \cdot \frac{\text{Var}(A')}{c^3} = 0 \\
c^3 &= 2\sqrt{G - 1} \cdot \text{Var}(A') \\
c^* &= \left[2\sqrt{G - 1} \cdot \text{Var}(A')\right]^{1/3}
\end{align}

Thus
\begin{align}
\|\nabla_\vtheta J_{\text{GRPO}}(\vtheta)\| \leq 3\cdot 2^{\frac{1}{3}} \cdot \gamma(x;\vtheta) \cdot (\sqrt{G - 1})^{1/3} \cdot (\text{Var}(A'))^{1/3}
\end{align}
\end{proof}

\subsection{Proof of Lemma~\ref{Lemma:Max_Var}}
\label{Section:Proof_Lemma_1}
\begin{proof}[Proof of Lemma~\ref{Lemma:Max_Var}]

We proceed by backward induction on $N$, starting from $N = M$ down to $N = 2$.

\underline{Base case ($N = M$):} Take $A' = A$. Then $\operatorname{Var}(A') = 1 \ge 1$.

\underline{Inductive step:}  
Assume for some $n+1$ with $2 \le n+1 \le M$ that there exists a subset $A'_{n+1} \subseteq A$ with $|A'_{n+1}| = n+1$ and $\operatorname{Var}(A'_{n+1}) \ge 1$.  
We will show that there exists a subset $A'_{n} \subseteq A'_{n+1}$ with $|A'_{n}| = n$ and $\operatorname{Var}(A'_{n}) \ge 1$ by removing an element $a \in A'_{n+1}$ to form $A'_{n} = A'_{n+1} \setminus \{a\}$. 

Considering definition of variance: 
\begin{align}
    n\operatorname{Var}(A'_{n})+(a-\mu_{n+1})^2 = (n+1)\operatorname{Var}(A'_{n+1})
\end{align}

We need $\operatorname{Var}(A'_{n}) \ge 1$, thus:
\begin{align}
    \frac{n+1}{n}\operatorname{Var}(A'_{n+1}) - \frac{(a-\mu_{n+1})^2}{n} \ge 1
\end{align}

Thus we need to find an $a \in A'_{n+1}$ that
\begin{align}
\label{Equation:Find_A}
    (a-\mu_{n+1})^2 &\le (n+1)\operatorname{Var}(A'_{n+1}) - n
\end{align}

When $\operatorname{Var}(A'_{n+1}) = 1$, it is easy to find an $a$ makes $(a-\mu_{n+1})^2 \le (n+1) - n = 1 = \operatorname{Var}(A'_{n+1})$ hold since $\operatorname{Var}(A'_{n+1})$ is the average of $(a - \mu_{n+1})^2$, the distance of elements to average, over $A'_{n+1}$.

When $\operatorname{Var}(A'_{n+1}) > 1$, we assume there is no $a$ makes Equation~\ref{Equation:Find_A} hold. Formally:
\begin{align}
\label{Equation:Assumption_no_A}
    \forall a \in A'_{n+1}, (a - \mu_{n+1})^2 > (n+1)\operatorname{Var}(A'_{n+1}) - n
\end{align}

Thus
\begin{align}
    \operatorname{Var}(A'_{n+1}) &> (n+1)\operatorname{Var}(A'_{n+1}) - n \\
    1 &> \operatorname{Var}(A'_{n+1})
\end{align}

which is conflict with $\operatorname{Var}(A'_{n+1}) \ge 1$. 

So assumption~\ref{Equation:Assumption_no_A} does not hold, which means when $\operatorname{Var}(A'_{n+1}) > 1$:
\begin{align}
    \exists a\in A'_{n+1}, (a - \mu_{n+1})^2 \le (n+1)\operatorname{Var}(A'_{n+1}) - n
\end{align}

So Equation~\ref{Equation:Find_A} always holds when $\operatorname{Var}(A'_{n+1}) \ge 1$ and $\operatorname{Var}(A'_{n}) \ge 1$. 
By induction, Lemma~\ref{Lemma:Max_Var} holds for all $N$ with $2 \le N \le M$.

\end{proof}
\section{Training hyper-parameters} \label{Section:Training_hyper_parameters}
Table~\ref{Table:Model_hyper_parameters} and~\ref{Table:Algorithm_hyper_parameters} list hyper-parameters used in experiments.
\begin{table}[htbp]
\centering
\caption{Generation configuration of different base models according to their official release.}
\label{Table:Model_hyper_parameters}
\resizebox{0.8\textwidth}{!}{%
\begin{tabular}{ccccc}
\hline
\textbf{Model} & \textbf{Epoch} & \textbf{Temperature} & \textbf{Top-p} & \textbf{Learning rate} \\ \hline
Qwen2.5-Math-7B       & 1 & 1.0 & 0.9  & 5e$^{-7}$ \\
Qwen2.5-Math-1.5B       & 1 & 1.0 & 0.9  & 5e$^{-7}$ \\
Llama3.1-8B-Instruct  & 2 & 0.6 & 0.9  & 1e$^{-8}$ \\
OpenMath2-Llama3.1-8B & 2 & 0.7 & 0.95 & 2e$^{-7}$ \\ \hline
\end{tabular}
}
\end{table}

\begin{table}[htbp]
\centering
\caption{Hyper parameters used in finetuning.}
\label{Table:Algorithm_hyper_parameters}
\resizebox{0.6\textwidth}{!}{%
\begin{tabular}{clc}
\hline
\textbf{Parameter} & \multicolumn{1}{c}{\textbf{Description}} & \textbf{Value} \\ \hline
$G$                & Group size of GRPO and GSPO              & 32             \\
$\epsilon$         & Clip thereshold of importance ratio      & 0.2            \\
$N_{\text{init}}$  & Init size of D$^3$S selected responses   & 8              \\
$N_{\text{final}}$ & Final size of D$^3$S selected responses  & 32             \\
$K_{\text{init}}$  & Init ratio of D$^3$S selected tokens     & 5\%            \\
$K_{\text{final}}$ & Final ratio of D$^3$S selected tokens    & 20\%           \\ \hline
\end{tabular}%
}
\end{table}

\section{Detailed experiment results} \label{Section:Detailed_results}

\subsection{Analysis about experiment results of Llama Model}
\label{Section:Ablation_Llama}
In our ablation studies, a noteworthy phenomenon emerged when applying the D$^3$S framework to the general-purpose Llama3.1-8B-Instruct model, which has not been specifically aligned for mathematical reasoning. As detailed in Table~\ref{Table:Ablation_Llama}, strategies based on intra-group sampling (D$^1$S and D$^3$S-I) demonstrated markedly superior performance compared to their cross-group counterparts (D$^1$S-C and D$^3$S). Specifically, D$^3$S-I, which omits the cross-group sampling component, achieved an average Pass@1/Pass@8 score of 24.0/40.2, surpassing the full D$^3$S configuration.

\begin{table}[htbp]
\centering
\caption{Ablation study by incrementally applying different part of strategies of D$^3$S to Llama3.1-8B-Instruct and OpenMath2-Llama3.1-8B. Performance across benchmarks measured in pass@1/pass@8. }
\label{Table:Ablation_Llama}
\resizebox{0.8\textwidth}{!}{%
\begin{tabular}{lcccccccc}
\hline
Model    & AIME24   & AMIE25           & AMC23              & GSM8k     & MATH      & Minerva   & Olympiad & \textbf{Average} \\ \hline
\multicolumn{9}{c}{Llama3.1-8B-Instruct}                                                                                      \\ \hline
base     & 1.7/10.9 & 0.4/2.8          & 15.0/47.3          & 57.7/92.8 & 29.3/55.9 & 14.7/38.5 & 2.2/8.2  & 17.3/36.6        \\
GRPO     & 2.0/5.0  & 0.0/0.0          & 13.7/33.4          & 78.6/93.5 & 31.5/52.0 & 15.9/35.6 & 2.1/7.2  & 20.5/32.4        \\
D$^1$S   & 4.1/14.6 & \textbf{0.7/5.1} & 21.8/56.8          & 76.9/94.4 & \textbf{37.5}/59.6 & 19.9/41.9 & \textbf{3.8/11.6} & 23.5/\textbf{40.6}        \\
D$^1$S-C & 2.7/9.3  & 0.1/0.8          & 14.4/35.7          & 78.4/93.4 & 31.3/51.7 & 16.2/35.5 & 1.9/7.0  & 20.7/33.3        \\
D$^2$S   & 1.9/6.5  & 0.1/1.0          & 13.1/32.3          & 77.6/93.5 & 32.3/53.0 & 15.9/36.8 & 2.3/7.7  & 20.5/33.0        \\
D$^3$S & \textbf{5.3/20.7} & 0.1/0.8           & 20.3/50.8          & \textbf{79.0/95.0} & 35.9/59.2 & 22.5/44.3 & 3.3/10.7 & 23.8/40.2          \\
D$^3$S-I & 4.4/18.8 & 0.5/4.2          & \textbf{23.0/57.8} & 77.8/95.0 & 36.2/\textbf{59.6} & \textbf{22.8/44.6} & 3.3/10.6 & \textbf{24.0}/40.2        \\ \hline
\multicolumn{9}{c}{OpenMath2-Llama3.1-8B}                                                                                     \\ \hline
base     & 3.3/12.2 & 2.0/10.2         & 35.5/60.7          & 89.1/96.2 & 49.8/65.6 & 11.8/26.4 & 7.8/15.0 & 28.5/40.9        \\
GRPO     & 5.8/17.8 & 2.0/10.0         & 35.8/59.9          & 85.7/95.6 & 50.7/65.3 & 10.0/22.9 & 7.4/14.0 & 28.2/40.8        \\
D$^1$S & \textbf{6.9/20.7} & \textbf{3.1/16.2} & \textbf{40.7/67.5} & \textbf{90.5}/96.1          & 52.6/66.2 & \textbf{14.0}/27.4 & 8.9/16.0 & \textbf{31.0/44.3} \\
D$^1$S-C & 6.8/20   & 2.5/11.6         & 35.9/63.6          & 89/95.9   & 52.2/\textbf{66.5} & 9.7/21.0  & 7.1/13.0 & 29.0/41.7        \\
D$^2$S   & 5.5/18.4 & 3/12.9           & 35.9/60            & 88.7/96   & 51.1/65.7 & 9.9/21.8  & 7.4/14.1 & 28.8/41.3        \\
D$^3$S   & 5.6/18.6 & 2.0/9.0          & 35.2/58.6          & 89.4/\textbf{96.2} & 49.6/64.9 & 11.3/26.3 & 8.1/15.2 & 28.7/41.3        \\
D$^3$S-I & 6.7/19.7 & 3.0/13.5         & 39.1/65.6          & 90.2/\textbf{96.2} & \textbf{52.7}/66.4 & 13.5/\textbf{27.5} & \textbf{9.0/16.4} & 30.6/43.6        \\ \hline
\end{tabular}%
}
\end{table}

\begin{figure}[htbp]
\centering
\begin{subfigure}[b]{0.32\textwidth}
\includegraphics[width=\textwidth]{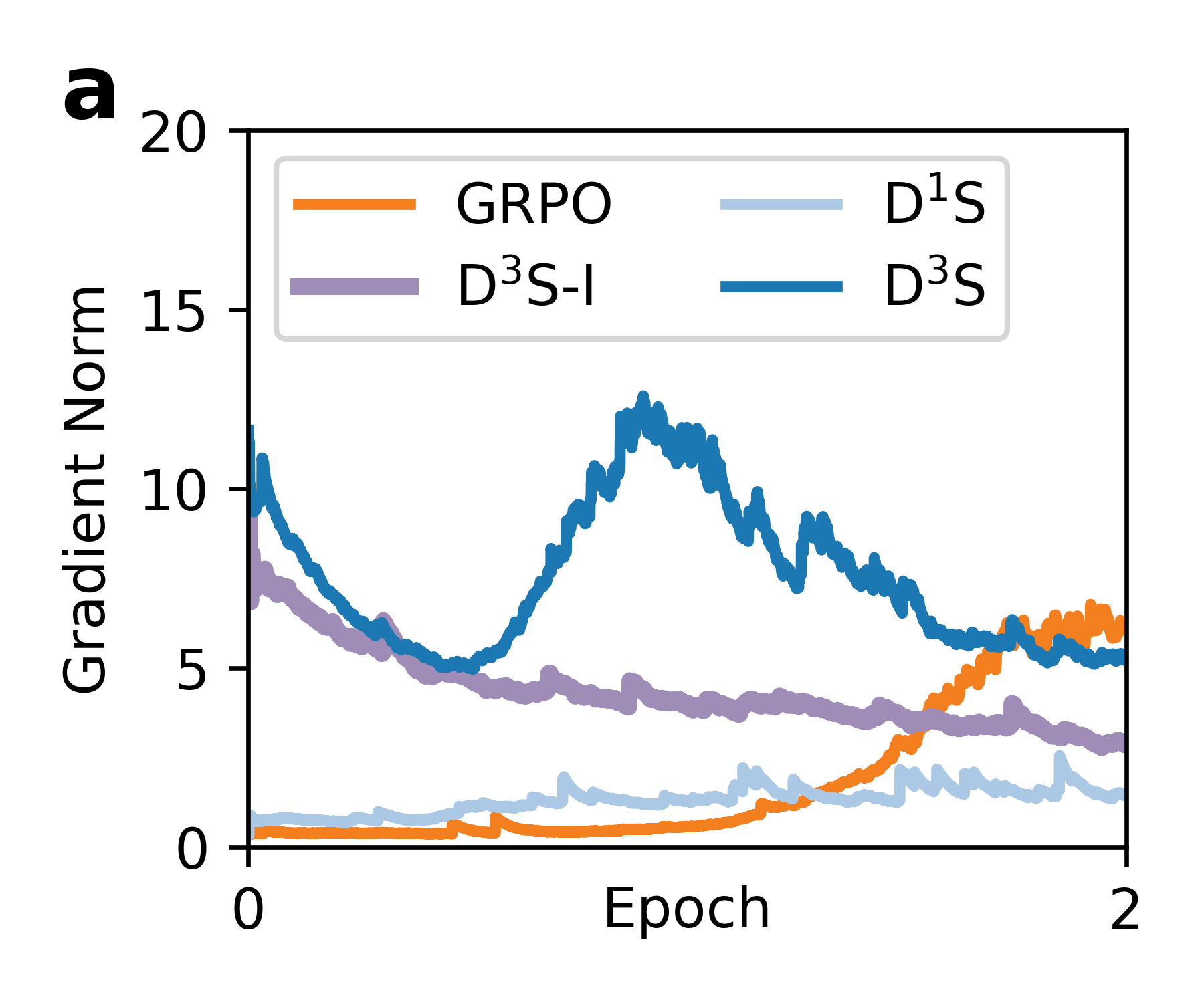}
\caption{}
\label{Figure:Train_Ablation_Grad_Norm_Llama_ema}
\end{subfigure}
\hfill
\begin{subfigure}[b]{0.32\textwidth}
\includegraphics[width=\textwidth]{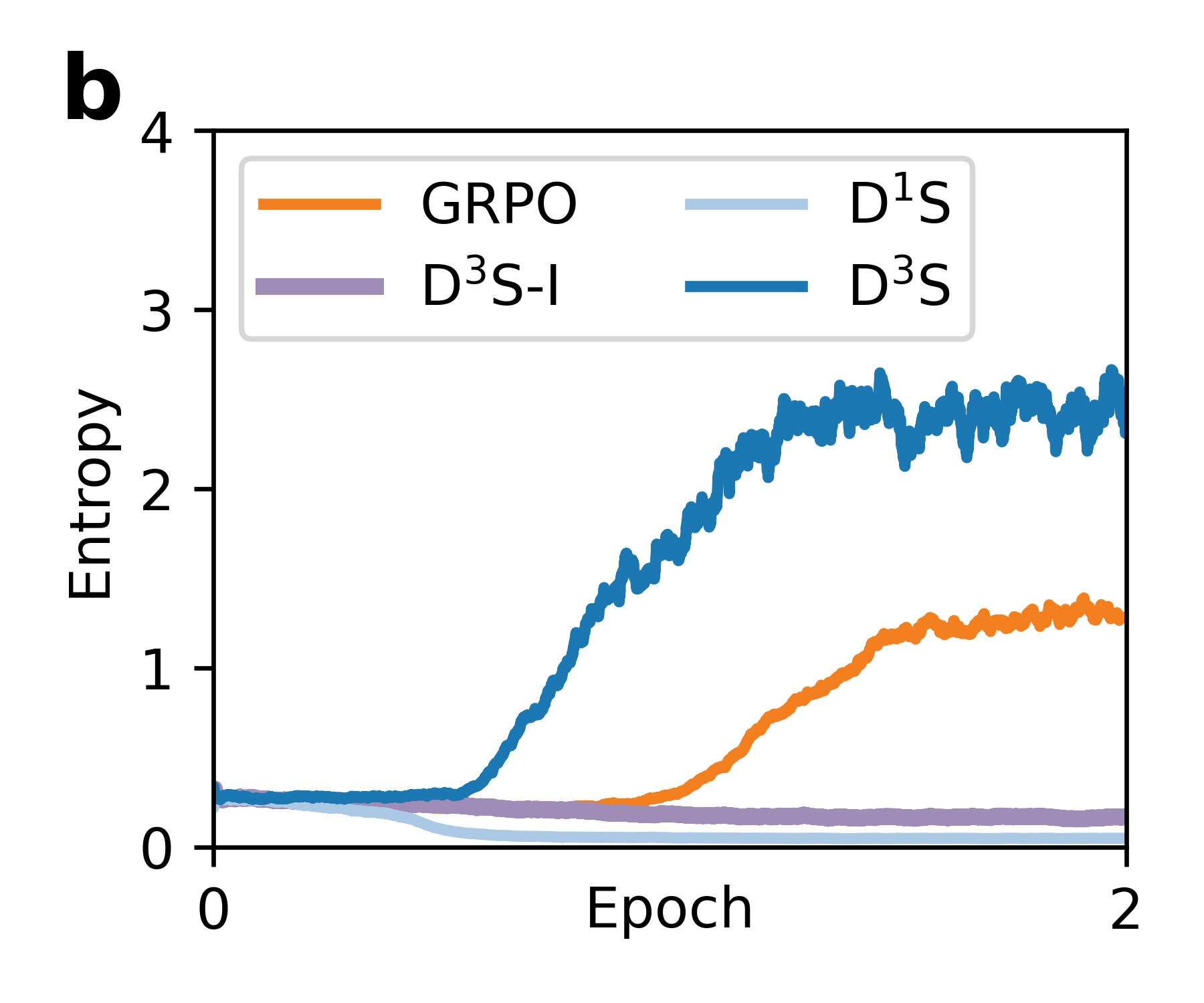}
\caption{}
\label{Figure:Train_Ablation_Entropy_Llama_ema}
\end{subfigure}
\hfill
\begin{subfigure}[b]{0.32\textwidth}
\includegraphics[width=\textwidth]
{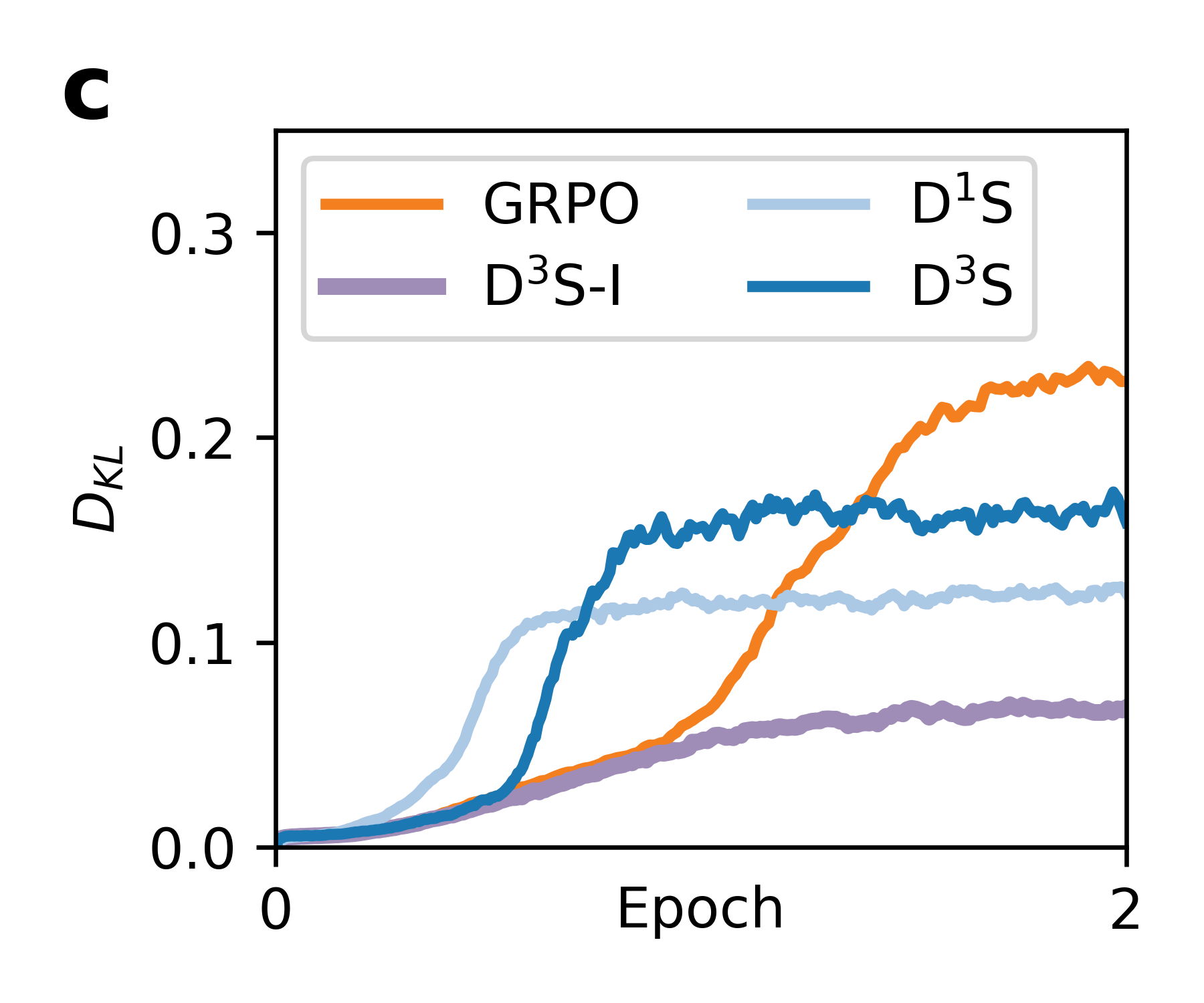}
\caption{}
\label{Figure:Train_Ablation_KL_Llama_ema}
\end{subfigure}
\caption{Detailed training dynamics of D$^3$S strategies with Llama3.1-8B-Instruct as base model, where D$^3$S-I denotes D$^3$S without cross-group down-sampling strategy. (\subref{Figure:Train_Ablation_Grad_Norm_Llama_ema}) gradient norms, (\subref{Figure:Train_Ablation_Entropy_Llama_ema}) policy entropy, and (\subref{Figure:Train_Ablation_KL_Llama_ema}) $\KL$ restrain. Compared to the original GRPO, the integration of D$^3$S significantly increases norm of policy gradient. Since Llama3.1-8B-Instruct model lacks pre-alignment, D$^3$S-I provides better balance in exploitation and exploration through smoother policy gradient and milder policy distribution migration measured in $\KL$.}
\label{Figure:Ablation_Llama}
\end{figure}

The training dynamics, depicted in Figure~\ref{Figure:Ablation_Llama}, provide a clear explanation for this discrepancy. The cross-group D$^3$S strategy (dark blue line) induced extremely high-variance and unstable policy gradients, as shown in Figure~\ref{Figure:Train_Ablation_Grad_Norm_Llama_ema}, accompanied by a sharp increase in $\KL$ (Figure~\ref{Figure:Train_Ablation_KL_Llama_ema}). This suggests that for an unaligned model with highly heterogeneous output quality across different prompts, the global selection mechanism over-concentrates the learning signal on a few outlier samples, leading to an unstable optimization process. In contrast, D$^3$S-I (purple line) maintained a policy gradient that was both high in magnitude and remarkably stable, with a much milder policy distribution migration. This indicates that for unaligned models, providing a stable, localized learning signal within each prompt's context is more effective than pursuing the globally maximal signal.

This behavior changes when the framework is applied to OpenMath2-Llama3.1-8B, a domain-aligned version of the same base model. The results in Table~\ref{Table:Ablation_Llama} show that the performance gap between intra-group and cross-group sampling strategies narrows considerably. While the intra-group variants D$^1$S (31.0/44.3) and D$^3$S-I (30.6/43.6) remain among the top performers, highlighting their robustness, the cross-group methods also yield competitive results. This suggests that as the model becomes better aligned and its response quality more consistent, the risk of instability from cross-group sampling diminishes, allowing its benefits to be more effectively realized.

This comparative analysis strongly indicates that the model's degree of alignment is a critical variable in determining the optimal sampling strategy. In this context, the D$^3$S-I variant stands out as a particularly robust framework. By combining the stability of intra-group sampling with the precision of token-level selection and dynamic scheduling, it delivers excellent performance and stable training dynamics across models with varying levels of initial capability, making it a more universally applicable solution.

\subsection{Pass@16 metrics}
\label{Section:Pass_16}
\begin{table}[htbp]
\centering
\caption{Performance across benchmarks measured in pass@16 calculated from 32 parallel runs.}
\label{Table:Pass_16}
\resizebox{0.8\textwidth}{!}{%
\begin{tabular}{lcccccccc}
\hline
Model           & AIME24        & AMIE25        & AMC23         & GSM8k         & MATH          & Minerva       & Olympiad      & \textbf{Average} \\ \hline
\multicolumn{9}{c}{Qwen2.5-Math-7B}                                                                                                                \\ \hline
base            & 42.8          & 19.4          & 82.4          & 92.2          & 71.1          & 42.7          & 18.2          & 52.7             \\
GRPO            & 42.7          & 28.0          & 89.0          & 96.3          & 73.1          & 50.2          & 22.3          & 57.4             \\
PODS            & 48.1          & 30.8          & 85.1          & 96.4          & 73.5          & 51.9          & 23.2          & 58.4             \\
D$^1$S          & 52.3          & 27.4          & 89.9          & 96.5          & 73.2          & 52.7          & 22.8          & 59.3             \\
D$^1$S-C        & 45.3          & \textbf{32.1} & 88.3          & 96.7          & 73.4          & 51.6          & 23.1          & 58.6             \\
D$^2$S          & 48.8          & 27.0          & 87.3          & 96.4          & 73.3          & 52.0          & 22.7          & 58.2             \\
\textbf{D$^3$S} & \textbf{54.6} & 31.6          & \textbf{91.2} & \textbf{96.9} & \textbf{73.9} & \textbf{53.4} & \textbf{23.3} & \textbf{60.7}    \\ \hline
\multicolumn{9}{c}{Llama3.1-8B-Instruct}                                                                                                           \\ \hline
base            & 17.9          & 4.6           & 59.8          & 95.3          & 62.0          & 45.1          & 11.0          & 42.2             \\
GRPO            & 6.7           & 0.0           & 44.7          & 94.9          & 57.2          & 41.2          & 9.7           & 36.3             \\
PODS            & 14.7          & 5.0           & 51.0          & 94.9          & 57.8          & 44.5          & 10.0          & 39.7             \\
D$^1$S          & 18.5          & \textbf{8.8}  & \textbf{70.0} & 95.6          & 64.7          & 47.4          & \textbf{14.6} & 45.7             \\
D$^1$S-C        & 13.8          & 1.7           & 47.4          & 95.0          & 56.8          & 40.5          & 9.4           & 37.8             \\
D$^2$S          & 9.4           & 2.0           & 42.2          & 95.1          & 58.1          & 43.0          & 10.2          & 37.1             \\
\textbf{D$^3$S} & \textbf{28.4} & 1.7           & 59.9          & \textbf{96.3} & \textbf{64.7} & \textbf{49.9} & 13.6          & \textbf{44.9}    \\ \hline
\multicolumn{9}{c}{OpenMath2-Llama3.1-8B}                                                                                                          \\ \hline
base            & 15.0          & 14.5          & 68.6          & \textbf{97.0} & 68.7          & 30.9          & 17.0          & 40.9             \\
GRPO            & 22.5          & 13.7          & 65.2          & 96.4          & 68.2          & 27.2          & 15.9          & 44.2             \\
PODS            & 22.6          & 17.7          & 67.9          & 96.7          & 68.3          & 25.8          & 16.2          & 45.0             \\
\textbf{D$^1$S} & \textbf{26.9} & \textbf{24.7} & \textbf{74.1} & 96.8          & \textbf{69.0} & \textbf{31.3} & \textbf{18.2} & \textbf{48.7}    \\
D$^1$S-C        & 24.2          & 15.7          & 70.2          & 96.8          & 69.6          & 24.5          & 14.6          & 45.1             \\
D$^2$S          & 25.9          & 18.1          & 67.6          & 96.7          & 68.8          & 25.8          & 15.9          & 45.5             \\
D$^3$S          & 24.0          & 13.2          & 65.3          & 96.9          & 68.1          & \textbf{31.3} & 17.0          & 45.1             \\ \hline
\end{tabular}%
}
\end{table}

\end{document}